# Underwater Acoustic Source Seeking Using Time-Difference-of-Arrival Measurements

Filip Mandić, *Member, IEEE*, Nikola Mišković, *Senior Member, IEEE*, and Ivan Lončar, *Student Member, IEEE*

***Abstract*—The research presented in this paper is aimed at developing a control algorithm for an autonomous surface system carrying a two-sensor array consisting of two acoustic receivers, capable of measuring the time-difference-of-arrival (TDOA) of a quasiperiodic underwater acoustic signal and utilizing this value to steer the system toward the acoustic source in the horizontal plane. Stability properties of the proposed algorithm are analyzed using the Lie bracket approximation technique. Furthermore, simulation results are presented, where particular attention is given to the relationship between the time difference of arrival measurement noise and the sensor baseline—the distance between the two acoustic receivers. Also, the influence of a constant disturbance caused by sea currents is considered. Finally, experimental results in which the algorithm was deployed on two autonomous surface vehicles, each equipped with a single acoustic receiver, are presented. The algorithm successfully steers the vehicle formation toward the acoustic source, despite the measurement noise and intermittent measurements, thus showing the feasibility of the proposed algorithm in real-life conditions.**

## I. Introduction

DETERMINING the location of an unknown acoustic signal in an underwater environment is of great interest to the scientific community, since acoustic pingers of various frequencies and signal strengths are commonly used for a variety of purposes, from marking the objects of interest such as black boxes (for search and rescue operations [1]), to tagging animals residing underwater (for marine biology purposes [2]), or simply marking the specific locations where an artifact or particular phenomena have been detected by the human divers (for marine archeology [3]).

In an underwater environment, global navigation satellite system signals, that are available and widely used for localization in numerous land and air applications, are absent due to the very weak propagation of the electromagnetic waves through the water. An electromagnetic homing system, presented in [4], is an example of the alternative that can provide an accurate measurement of the autonomous vehicle position and orientation to the dock during the homing, but also with very limited range due to the same propagation constraints. Therefore, in the underwater environment, the acoustic-based localization techniques are predominantly used. The conventional methods for locating underwater acoustic signals include using ultrashort baseline systems that usually rely on the two-way acoustic signal travel time for determining precise range, bearing, and elevation of the object of interest. Also, single range localization is used as an underwater environment localization technique [5]–[7], where range measurements are acquired using the acoustic modem devices. When employing these devices, range can be measured using the one-way travel time (OWTT) or two-way travel time (TWTT) ranging techniques. OWTT range is estimated from the time-of-flight (TOF) of acoustic data packets propagating between the underwater acoustic modems on the source and the vehicle side. This kind of ranging requires precision clocks to synchronize the modems. One-way TOF is then calculated from the difference between the time-of-launch (TOL) encoded in the acoustic packet, measured by a precision clock aboard the source, and the time-of-arrival (TOA), measured by a precision clock aboard the underwater vehicle. The accuracy of these TOF measurements is limited by the accuracy of the installed precision clocks. To ensure valid TOF measurements, it is crucial that the clocks on the sender and the receiver side remain synchronized throughout the dive to within an acceptable tolerance, which is difficult to achieve in practice over the longer periods of time [8]. TWTT is the most commonly used ranging technique. It requires the interaction between the source and the vehicle in such a way that the vehicle side acoustic modem sends a request, marking TOL, to the source modem which then responds to the request. The vehicle modem receiving a reply from the source marks TOA, and the range is then estimated from the calculated time difference. Since TOL and TOA are both measured on the vehicle side, TWTT does not require a clock synchronization between the two acoustic devices.

The problem of reaching a source signal, formally known as the source seeking problem, is quite commonly addressed in the recent scientific literature. The presented objective is determining the minimum or maximum of an unknown signal field. The majority of interest in the area stems from the need to achieve the full vehicle autonomy in an unstructured environment where

Manuscript received May 9, 2018; revised September 25, 2018; accepted January 21, 2019. This work was supported in part by the Croatian Science Foundation under Project IP-2016-06-2082 "Cooperative robotics in marine monitoring and exploration" and in part by the European Union through the EU H2020 FET-Proactive project "subCULTron," No. 640967. The work of F. Mandić was supported by the Croatian Science Foundation through the Project for the young researcher career development No. I-3485-2014. *(Corresponding author: Filip Mandić.)*

**Associate Editor: B. Englot.**

The authors are with the University of Zagreb, Faculty of Electrical Engineering and Computing, Laboratory for Underwater Systems and Technologies, Zagreb 10000, Croatia (e-mail: filip.mandic@fer.hr; nikola.miskovic@fer.hr; ivan.loncar@fer.hr).

the position of a specific object of interest must be reached, but, due to a variety reasons, no position measurements are available. Such conditions can be found in different applications, e.g., the pollutant source detection [9] or an autonomous vehicle homing [10]. Usually, the problem being considered is that of seeking the source of a scalar signal that decays away from the source, e.g., [11] and [12]. Such signal can be electromagnetic, acoustic, thermal, or a concentration of a chemical or biological agent. In [13], a classical extremum seeking scheme for navigating a stable and moderately unstable force-actuated point mass in a 2-D plane was presented. In [12], an extremum seeking scheme for systems with slow or drifting sensors was proposed, whereas in [14], a form of extremum seeking that guarantees known bounds on the update rates and control efforts was introduced. An extremum seeking scheme is usually deployed when the system model is not known very well or even remains completely unknown. A significant advantage of the extremum seeking is the fact that the extremum seeking control loop compensates the constant disturbances acting on the vehicle, such as gravity, buoyancy, or currents.

Source seeking with movement toward the object of interest in the underwater environment can be achieved using the range measurements, e.g., acoustic range measurements are used in [15], where the extremum seeking-based control approach from [16] is presented as a means to converge toward an underwater source, whereas in [5] an extended Kalman filter is used to determine a source location using the range measurements, and the vehicle's conventional control algorithms are then used to reach the desired position. One of the main issues with estimation-based single range localization systems is the observability of the system. There is a great number of papers dealing with observability of the range-only navigation systems using the different methodologies (see [6], [17], and [18]). Should the system not be observable, range measurements can result in false navigation. Therefore, to estimate the source position, the vehicle that is navigating using the range measurements from a stationary source needs to preform trajectories along which the system is observable. In [10], a method for homing an autonomous underwater vehicle to a subsea docking station is presented, where a sum of Gaussian filter is used to estimate the docking location while the autonomous vehicle is guided along an observable trajectory.

As already mentioned, to acoustically acquire range measurements, a cooperation with the underwater acoustic source (in the form of a send–reply communication, or clock synchronization) needs to be established, making these systems inapplicable to solving the problem of determining the location of an unknown underwater acoustic source. One solution to this problem can be found in applying the time-difference-of-arrival (TDOA) techniques, which require having at least three receivers that allow the localization of the unknown acoustic source in the horizontal plane [19]. In general, TDOA is a method that calculates the source location from the differences of arrival times measured on transmission paths between the source and fixed receivers [20]. The TDOA-based localization schemes usually consist of two steps: the measurement acquisition step and the multilateration step. In the measurement acquisition step, the differences of acoustic signal arrival times on several receiver nodes are measured. Based on the property of a hyperbola, the source will be located on a hyperbola whose difference between the ranges to the respective receivers is a constant. The difference in ranges is easily calculated from the measured difference in TOA and the known speed of the acoustic signal. With more than two receivers, more hyperbolic functions can be computed, which ideally intersect in one unique point, thus determining the source location [21]. Traditional 3-D underwater localization techniques require four noncoplanar receivers to localize the underwater signal source successfully [22]. However, that need can be eliminated via the use of depth information acquired by a pressure sensor, and a projection-based technique that translates the receiver nodes to the plane of the signal source [19]. This makes the localization using only three anchors possible, assuming the projection of the three noncollinear anchors is nondegenerative.

The research presented in this paper is aimed at developing a control algorithm for an autonomous surface system capable of measuring the TDOA of a quasiperiodic underwater acoustic signal passively emitted on a predefined frequency by an acoustic pinger, then utilizing this value to steer the system toward the signal source in the horizontal plane. In the scenario presented herein, the TDOA is measured only for one pair of acoustic receivers. Initially, at the TOL, the acoustic signal is emitted from the acoustic source. The signal then travels through the acoustic medium, and is received by the two acoustic receivers at their respective TOA instants. The assumption is that the TOL is unknown, and the quasiperiodicity in the envisioned scenario means that the acoustic source regularly emits the signal but not necessarily with an identical period each time. To the authors' best knowledge, the approach presented herein differs from the other source seeking approaches in that it does not use a signal amplitude to determine source location. Instead, a single TDOA measurement, acquired by adding a second receiver, coupled with an appropriate receiver movement strategy, enables convergence to the signal source in the horizontal plane. One of the advantages of the proposed method over the methods that use the signal amplitude to determine source location is that it can be unambiguously determined when the source has been reached, since in that case for any change of the baseline orientation, that is formed by two receivers, measured TDOA remains zero. When the signal amplitude is used that is not the case, even when the maximum amplitude of the signal is known, since the depth to the source is unknown. Also, much less deviation form the shortest path between the source and the vehicle is needed during the convergence phase to the source compared to a perturbation-based source seeking systems using signal amplitude. Approaches using two moving receivers can also be found in [23] and [24]. These approaches are estimation-based, unlike the approach presented herein, where raw measurements are used as a control algorithm input. Therefore, the main contributions of the paper are as follows:

1) development of a source seeking control algorithm that enables convergence toward an underwater acoustic signal source in the horizontal plane by using the single TDOA measurements;
2) stability analysis of the proposed algorithm;
3) experimental validation of the proposed algorithm.

This paper is organized as follows. Section II presents the mathematical model of the problem and introduces surge speed and yaw rate controllers deployed in the source seeking scheme. In the Section III, stability analysis of the proposed control algorithm using the Lie bracket approximation technique is given. Section IV describes the practical implementation limitations, discusses how to mitigate them and presents the complete algorithm implementation scheme. The simulation results of the proposed control scheme are given in Section V where influence of the measurement noise on algorithm operation is the main focus. Section VI presents the experimental results obtained in a real-life environment. This paper is concluded with Section VII.

## II. Problem Description and Control Algorithm

In this section, the source seeking problem using TDOA measurements is described and the system's kinematic model is given. Then, the control algorithm, consisting of a surge speed and yaw rate controllers that steer the system toward the source, is presented.

### A. Problem Description

The system holding two acoustic receivers is modeled as a nonholonomic unicycle, where the heading angle in the global coordinate frame $\{N\}$ is denoted with $\psi$, and the position of the vehicle center is denoted with $\mathbf{p_c} = [x_c \;\; y_c \;\; 0]^T$. The acoustic sensors are mounted orthogonally respective to the system heading $\psi$, thus forming a baseline with point $\mathbf{p_c}$ in the middle of the baseline. The acoustic source position is denoted as $\mathbf{p_t} = [x_t \;\; y_t \;\; z_t]^T$, and the position of the acoustic receivers as $\mathbf{p_i} = [x_i \;\; y_i \;\; 0]^T$ where $i = \{1, 2\}$. Therefore, the acoustic receiver positions are defined with $\mathbf{p_1} = \mathbf{p_c} + \mathbf{p_{off}}$ and $\mathbf{p_2} = \mathbf{p_c} - \mathbf{p_{off}}$ where vector $\mathbf{p_{off}} = [(d/2)\sin\psi \;\; -(d/2)\cos\psi]^T$ is the offset of the receivers from the baseline center $\mathbf{p_c}$ in their respective directions. The parameter $d$ denotes the length of the baseline formed by the acoustic receivers. Depending on the application, if the sufficient baseline can be achieved, the acoustic receivers can be mounted on a single vehicle, or on two separate vehicles keeping a formation that allows a variable baseline length. If mounted on a single vehicle, independent control of the system yaw and surge degrees of freedom is required for the algorithm operation, while in the scenario with two vehicles, it is required that the vehicles are fully actuated to keep the formation. However, in Section IV-B, it is discussed that even in the single vehicle scenario it is desirable that the vehicle be fully actuated to compensate for the disturbances affecting it. In the scenario where the vehicles are keeping a formation, it is further assumed that a formation-keeping algorithm is deployed on the controlled vehicles, so that, for the purpose of this analysis, we can mathematically observe both the formation and the vehicle as a single entity with virtual actuators that are used to impart the forward velocity $u_c$ and the angular velocity $\dot{\psi}$.

Only information about the source position can be inferred from the TDOA measurement, therefore, using the notation introduced above, the signal TOA difference $\Delta_{\mathrm{TOA}}$ is defined as

$$\Delta_{\mathrm{TOA}} = \mathrm{TOA}_1 - \mathrm{TOA}_2 = \frac{\|\mathbf{p_1} - \mathbf{p_t}\| - \|\mathbf{p_2} - \mathbf{p_t}\|}{v} \tag{1}$$

where $\mathrm{TOA}_1$ and $\mathrm{TOA}_2$ present the TOA measurements taken at the respective receivers and $v$ is the speed of sound in the water. Going further, a normalized nondimensional measurement $\Delta \in [-1, 1]$, given in the form

$$\Delta = \Delta_{\mathrm{TOA}} \frac{v}{d} \tag{2}$$

is used in the control algorithm. The overall control algorithm goal—the baseline center convergence toward the acoustic source in the horizontal plane—is achieved through two subgoals: first, to orient the baseline toward the signal source, and second, to approach the signal source. These subgoals are achieved through yaw rate and surge speed control of the system, respectively.

### B. Yaw Rate Control Law

The first subgoal, positioning of the acoustic receiver baseline orthogonally with respect to the bearing vector between the source and the baseline center, can be achieved by bringing the $\Delta$ value to zero. To do that, in the yaw degree of freedom (5), a gradient-based extremum seeking controller, as presented in [16], is employed. The Cartesian representation of the unicycle system with such control of the yaw degree of freedom is described with the following set of differential equations:

$$\dot{x}_c = u_c \cos\psi \tag{3}$$

$$\dot{y}_c = u_c \sin\psi \tag{4}$$

$$\dot{\psi} = k(f - x_e h)\sqrt{\omega}\sin\omega t + a\sqrt{\omega}\cos\omega t \tag{5}$$

$$\dot{x}_e = -x_e h + f. \tag{6}$$

The states $x_c$ and $y_c$ in (3) and (4) are kinematic equations of a unicycle-like model [25]. Equation (5) defines the extremum seeking control law in the yaw degree of freedom where a sinusoidal perturbation $a\sqrt{\omega}\cos\omega t$ with a perturbation frequency $\omega$ and a scaling factor $a$ is added to persistently excite the system while the corresponding demodulation term $\sqrt{\omega}\sin\omega t$ is used to estimate the gradient of the nonlinear map, i.e., the cost function $f$. The cost function that achieves the minimum for $\Delta = 0$ is defined as $f = \Delta^2$. The term $(f - x_e h)$ represents the highpass filter that is a part of the extremum seeking scheme and it is used to remove possible constant offsets present in the cost function measurement. The state $x_e$ in (6) is the filter state and the parameter $h$ is a filter time constant.

### C. Surge Speed Control Law

Since the extremum seeking control law in (5) only ensures that the baseline is orthogonal with respect to the bearing vector, an additional surge speed control law needs to be introduced. It should steer the center of the baseline toward the source of the acoustic signal, thus achieving the second subgoal. To derive an appropriate surge speed control law and later show the stability of the proposed control scheme we resort to use of polar coordinates, defined as

$$r_c = \sqrt{(x_c - x_t)^2 + (y_c - y_t)^2} \quad (7)$$

$$\theta_c = \text{atan2}(y_c - y_t, x_c - x_t) \quad (8)$$

where $r_c$ denotes the range and $\theta_c$ denotes the bearing between the baseline center and the source of the acoustic signal in the horizontal plane. Using the relations

$$\begin{bmatrix} x_c \\ y_c \end{bmatrix} = r_c \begin{bmatrix} \cos(\theta_c) \\ \sin(\theta_c) \end{bmatrix} + \begin{bmatrix} 1 & 0 & 0 \\ 0 & 1 & 0 \end{bmatrix} \mathbf{p_t}$$

$$\begin{bmatrix} \dot{x}_c \\ \dot{y}_c \end{bmatrix} = u_c \begin{bmatrix} \cos(\psi) \\ \sin(\psi) \end{bmatrix} \quad (9)$$

and assuming that the source is located at the origin of the coordinate frame, i.e., $\mathbf{p_t} = \begin{bmatrix} 0 & 0 & z_t \end{bmatrix}^T$, the differential equations in the polar coordinate system are derived as

$$\dot{r}_c = u_c \cos(\psi - \theta_c) \quad (10)$$

$$\dot{\theta}_c = \frac{1}{r_c} \left[ u_c \sin(\psi - \theta_c) \right]. \quad (11)$$

*Remark:* In the controlled system (3)–(6), the influence of the source depth $z_t$ is not explicitly present. However, its influence appears in the cost function $f$ amplitude. Since there is no movement in the vertical axis, we use the polar coordinates instead of cylindrical coordinates. Influence of the source depth on the algorithm performance is discussed in Sections III and V.

Normalized TOA difference measurement $\Delta$ in polar coordinates is given with $\Delta = (r_1 - r_2)/d$, where ranges between the acoustic receivers and the signal source are expressed with

$$r_1 = \sqrt{d^2/4 + r_c^2 + d r_c \sin(\psi - \theta_c) + z_t^2} \quad (12)$$

$$r_2 = \sqrt{d^2/4 + r_c^2 - d r_c \sin(\psi - \theta_c) + z_t^2}. \quad (13)$$

Finally, the polar representation of the system (3)–(6) is

$$\dot{r}_c = -u_c \cos\alpha \quad (14)$$

$$\dot{\theta}_c = -\frac{u_c}{r_c} \sin\alpha \quad (15)$$

$$\dot{\alpha} = k(f - x_e h)\sqrt{\omega} \sin\omega t + a\sqrt{\omega} \cos\omega t + \frac{u_c}{r_c} \sin\alpha \quad (16)$$

$$\dot{x}_e = -x_e h + f \quad (17)$$

where the substitution $\alpha = \psi - \theta_c + \pi$ is introduced. The state $\alpha$ represents the relative angle formed by the bearing vector between the source and the baseline center and baseline heading, as can be seen in Fig. 1.

After the polar coordinate transformation has been done, a surge speed control law can be introduced. The surge speed control law needs to accomplish three tasks. The first one is determining the surge speed sign, i.e., the direction of the signal source. From the single $\Delta$ value, and only if $\Delta$ equals zero, we can infer in which plane the acoustic signal source is located, or, looking merely at the horizontal plane, on which line the source point is located. However, the direction in which the system should be steered to reach the desired point is not known. The main idea is to exploit the perturbations introduced in the yaw degree of freedom to determine the desired direction in a fashion that somewhat resembles a sound localization present in human beings—in everyday perception, human beings use head motion to determine the sound direction and the distance in an attempt to acquire sound sources visually. Therefore, an important source of information for localization is obtained from the head motion cues. When we hear a sound we wish to localize, we move our head to minimize the interaural time differences, i.e., the arrival time of a sound between two ears, using our head as a sort of "pointer" [26].

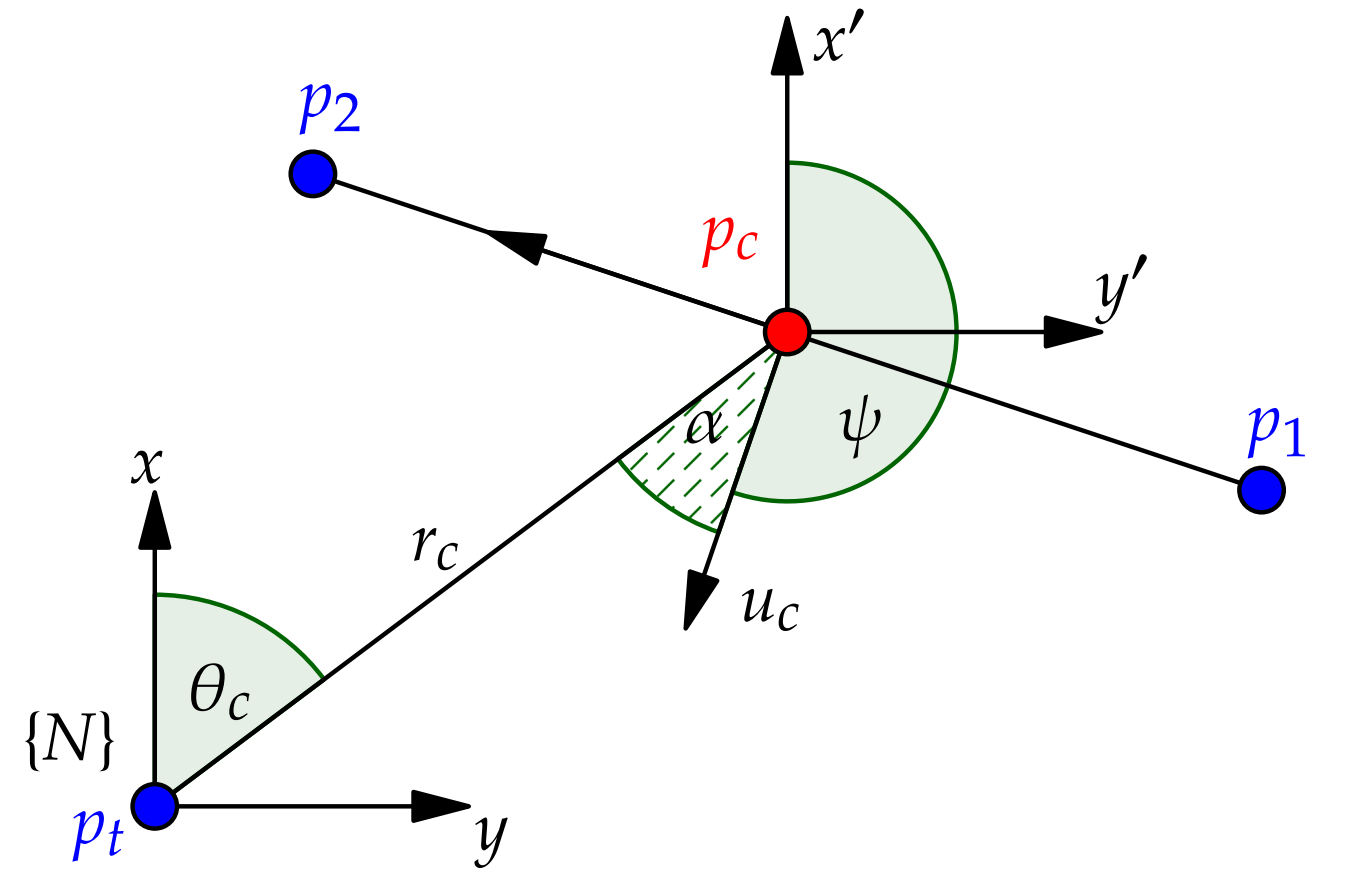

Fig. 1. Sensor baseline $d$ formed by the acoustic sensors located at $\mathbf{p_1}$ and $\mathbf{p_2}$, where $\mathbf{p_c}$ denotes the baseline center, and underwater acoustic signal source position $\mathbf{p_t}$ depicted in the horizontal plane.

The second task is to ensure that the system has surge speed greater than zero only when the cost function $f$ is low enough, meaning that the system is oriented toward the source. In this way, the vehicle will not diverge from the source while its heading is still converging under the influence of the extremum seeking controller.

The third task is reducing the surge speed to zero when the system has approached the target. When the system traveling at a constant surge speed passes over the target, the surge speed sign suddenly changes, and an oscillating behavior appears, meaning that the system position cannot ultimately settle even if it has reached the source. In addition, it quickly overshoots the source and has to go backwards if it was traveling at full speed in the vicinity of the source.

Finally, the surge speed control law that accomplishes all stated tasks is introduced as

$$u_c = u_\zeta \underbrace{\text{sgn}(\Delta \sin(2\alpha))}_{u_{\text{dir}}} \cdot \underbrace{u_0 (1 - \tanh(mf))}_{u_{\text{amp}}}. \quad (18)$$

The part of the control law denoted with $u_{\text{amp}}$ handles the requirement regarding the surge speed amplitude, where parameter $u_0$ is the maximum surge speed the system can achieve and $m$ is a tunable parameter that determines how much the system heading can deviate from the signal source direction before the surge speed is set to zero. Part $u_{\text{dir}}$ handles the surge sign resolution. The state $\alpha$ constantly changes due to perturbations and both $\alpha$ and $\Delta$ being positive signifies that the signal source is in front of the vehicle, whereas, if their signs are opposite, the signal source is behind the vehicle. A nonlinear damping term $u_\zeta$ is added to tune the forward velocity in the vicinity of the target. The properties that $u_\zeta$ has to satisfy are

$$u_\zeta \in [0, 1), \ u_\zeta \approx 1 \text{ for } r_c \gg \varepsilon, \text{and} \lim_{r_c \to 0} u_\zeta = 0. \tag{19}$$

This means that $u_\zeta \approx 1$ for a range sufficiently larger than some user defined value $\varepsilon$ and, as range reduces, the term approaches zero, e.g., the following class of damping function $u_\zeta$ satisfies given conditions

$$u_\zeta = \frac{r_c^q}{(r_c + \varepsilon)^q}. \tag{20}$$

In Section I, we emphasized that in the envisioned scenario, the range measurements are not available, however in Section IV a signal that satisfies the conditions given in (19) will be presented. Furthermore, in a real-life situation, the relative angle $\alpha$ cannot be measured since the source position is not known. This problem is also considered in Section IV.

## III. Stability Analysis

Using the Lie bracket approximation results from [27], we investigate the stability of the system defined with (14), (16), and (17). The procedure starts by writing the extremum seeking system in an input-affine form, then calculating its corresponding Lie bracket system. From there, the asymptotic stability of the Lie bracket system, if proven, implies practical asymptotic stability for the initial extremum seeking system.

The following class of input-affine systems is considered

$$\dot{\boldsymbol{\eta}} = \boldsymbol{b_0}(t, \boldsymbol{\eta}) + \sum_{i=1}^{m} \boldsymbol{b_i}(t, \boldsymbol{\eta})\sqrt{w} g_i(t, wt) \tag{21}$$

with $\boldsymbol{\eta}(t_0) = \boldsymbol{\eta_0} \in \mathbb{R}^n$ and $\omega \in (0, \infty)$. The Lie bracket system corresponding to (21) is defined with

$$\dot{\boldsymbol{z}} = \boldsymbol{b_0}(t, z) + \sum_{\substack{i=1 \\ j=i+1}}^{m} \left[\boldsymbol{b_i}, \boldsymbol{b_j}\right](t, z)\nu_{ji}(t) \tag{22}$$

where

$$\nu_{ji}(t) = \frac{1}{T}\int_0^T g_j(t, \theta)\int_0^\theta g_i(t, \tau)d\tau d\theta. \tag{23}$$

To perform the Lie bracket averaging step, several assumptions on $\boldsymbol{b_i}$ and $g_i$ are imposed in [27] and herein a summary is given. First, the vector fields $\boldsymbol{b_i}$ are required to be $C^2$ smooth and expressions involving $\boldsymbol{b_i}$ and their derivatives must be bounded uniformly in $t$. The inputs $g_i$ must be measurable and for all $i = 1, \ldots, N$ there should exist constants $L_i, M_i \in (0, \infty)$ such that $|g_i(t_1, \theta) - g_i(t_2, \theta)| \leq L_i\,|t_1 - t_2|$ for all $t_1, t_2 \in \mathbb{R}$ and such that $\sup_{t,\theta \in \mathbb{R}} |g_i(t, \theta)| \leq M_i$. Furthermore, they have to be $T$-periodic, and have a zero average. In [27], the authors give and prove a theorem, which states that the semiglobal (local) practical uniform asymptotic stability of the input-affine system (21) follows from the global (local) uniform asymptotic stability of the corresponding Lie bracket system (22) if stated assumptions are satisfied.

We start the stability proof by rewriting the system (14), (16), and (17) in a form suitable for the Lie bracket approximation step

$$\dot{\boldsymbol{\eta}} = \begin{bmatrix} \dot{r}_c \\ \dot{\alpha} \\ \dot{x}_e \end{bmatrix} = \underbrace{\begin{bmatrix} -u_c \cos\alpha \\ \frac{u_c}{r_c}\sin\alpha \\ -x_e h + f \end{bmatrix}}_{\boldsymbol{b_0}} + \underbrace{\begin{bmatrix} 0 \\ k(f - x_e h) \\ 0 \end{bmatrix}}_{\boldsymbol{b_1}} \sqrt{\omega}\underbrace{\sin(\omega t)}_{g_1} + \underbrace{\begin{bmatrix} 0 \\ a \\ 0 \end{bmatrix}}_{\boldsymbol{b_2}} \sqrt{\omega}\underbrace{\cos(\omega t)}_{g_2}. \tag{24}$$

The sinusoidal inputs $g_1$ and $g_2$ used in the extremum seeking scheme satisfy the assumptions stated earlier, as shown in [27], while the validity of the assumptions on the vector fields $\boldsymbol{b_0}$, $\boldsymbol{b_1}$, and $\boldsymbol{b_2}$ is shown in the Appendix. The Lie bracket system that captures the behavior of the trajectories of the original extremum seeking system, computed from (24), is

$$\dot{\tilde{r}}_c = -u_c \cos\tilde{\alpha} \tag{25}$$

$$\dot{\tilde{\alpha}} = \frac{u_c}{\tilde{r}_c}\sin\tilde{\alpha} + \frac{1}{2}ak\frac{\partial f(\tilde{r}_c, \tilde{\alpha})}{\partial\tilde{\alpha}} \tag{26}$$

$$\dot{\tilde{x}}_e = -\tilde{x}_e h + f \tag{27}$$

where $(\partial f/\partial\tilde{\alpha})(\tilde{r}_c, \tilde{\alpha})$ is

$$\frac{\partial f(\tilde{r}_c, \tilde{\alpha})}{\partial\tilde{\alpha}} = \frac{\tilde{r}_c^2 \sin(2\tilde{\alpha})}{r_1 r_2}. \tag{28}$$

Next, we must show the asymptotic stability of the derived Lie bracket system. Going forward, the damping function (20) with $q = 3$ is used. Equilibrium states are $\boldsymbol{z_e} = [0 \;\; n\pi \;\; 0]^T$ where $n \in \mathbb{Z}$. The relative angle $\alpha$ state has multiple equilibrium points. Geometrically, in the case of an even $n$, the source is in front of the baseline, while in the case of an odd $n$, the source is behind the baseline. For an odd $n$, if we shift the state $\alpha$, the signs of the trigonometric functions change, i.e., $\sin(\tilde{\alpha} + n\pi) = -\sin\tilde{\alpha}$ and $\cos(\tilde{\alpha} + n\pi) = -\cos\tilde{\alpha}$, but at the same time the $u_{\text{dir}}$ term changes its sign, so in (25) and (26) they cancel each other out. With that in mind, we proceed by analyzing only the equilibrium point $\boldsymbol{z_e} = \boldsymbol{0}^T$ without losing generality. We define a positive definite Lyapunov function

$$V = V_{r_c} + V_\alpha + V_{x_e} = \frac{1}{2}\tilde{r}^2 + \frac{1}{2}\tilde{\alpha}^2 + \frac{1}{2}\tilde{x_e}^2. \tag{29}$$

By taking the derivative of the function $V$, we get

$$\dot{V}_{r_c} = -\tilde{r}_c u_c \cos\tilde{\alpha} \tag{30}$$

$$\dot{V}_\alpha = \tilde{\alpha}\frac{u_c}{\tilde{r}_c}\sin\tilde{\alpha} + \frac{1}{2}\tilde{\alpha}ak\frac{\tilde{r}_c^2}{r_1 r_2}\sin 2\tilde{\alpha} \tag{31}$$

$$\dot{V}_{x_e} = -\tilde{x}_e^2 h + \tilde{x}_e f(\tilde{r}_c, \tilde{\alpha}). \tag{32}$$

First, note that by selecting an appropriate value of the parameter $m$ in (18), the term $u_{\text{amp}}$ can be tuned in such way that for each $\tilde{\alpha} \notin (-\delta, \delta)$ where $|\delta| < (\pi/2)$ the surge speed value $u_c$ tends to zero. Also, note that the following is valid for each $\tilde{\alpha} \in (-(\pi/2), \pi/2)$

$$\cos(\tilde{\alpha}) > 0, \quad \tilde{\alpha}\sin(\tilde{\alpha}) \geq 0, \quad \tilde{\alpha}\sin(2\tilde{\alpha}) \geq 0. \tag{33}$$

When the angle $\alpha \in (-(\pi/2), -\delta) \cap (\delta, \pi/2)$, i.e., it is far away from the equilibrium value, the derivative (31) reduces to $(1/2)\tilde{\alpha}ak((d^2\tilde{r}_c^2)/(r_1r_2))\sin 2\tilde{\alpha}$ and from (33) it follows that $\dot{V}_\alpha \leq 0$ if parameter $k < 0$ for each $\alpha \in (-(\pi/2), \pi/2)$. Moreover, $\dot{V}_\alpha$ is negative definite since $\dot{V}_\alpha = 0$ only for $\tilde{\alpha} = 0$ when $\tilde{r}_c > 0$. Note, for $\tilde{r}_c = 0$, $\dot{V}_\alpha$ is also 0, which is acceptable since the baseline center is already at the desired position, and angle $\alpha$ is not defined. For angle $\tilde{\alpha} \in (-\delta, \delta)$, the surge speed $u_c$ is present, which can be seen as a disturbance for the yaw rate controller. The derivative (31) can be written as

$$\tilde{r}_c^2\tilde{\alpha}\sin(\tilde{\alpha})\left(\frac{u_{\text{dir}}u_{\text{amp}}}{(\tilde{r}_c+\varepsilon)^3} + \frac{ak}{r_1r_2}\cos(\tilde{\alpha})\right) \leq 0 \tag{34}$$

where the expression is negative definite if the sum inside the brackets is negative. For large range values, term $(r_c + \varepsilon)^3$ dominates and decreases the value of the positive term inside the bracket. Therefore, we observe the theoretical worst case scenario where $\tilde{r}_c \approx 0$ in which $\max(u_{\text{amp}}/(\tilde{r}_c + \varepsilon)^3) = u_0/\varepsilon^3$, $r_1r_2 \approx d^2/4 + z_t^2$ and $\cos(\tilde{\alpha}) = \cos(\delta)$. A short calculation yields the following condition

$$u_0\left(d^2 + 4z_t^2\right) + 4ak\varepsilon^3\cos(\delta) \leq 0. \tag{35}$$

The parameters $k$, $a$, $d$, $u_0$, and $\varepsilon$ are design parameters and, by appropriate tuning, the condition (35) can be satisfied. Looking at (31), we can observe how target depth influences the yaw degree of freedom convergence. It is assumed that depth is constant. When the target depth $z_t$ is very high, so is the product $r_1r_2$, meaning that the value of the second term in (31), which forces the state to converge, decreases. Although $z_t$ is unknown, in most use cases the maximum depth of an area can be found out in advance and parameters can be tuned accordingly to achieve the desired performance. Finally, checking the case when the initial condition is $\tilde{\alpha}_0 = \pm(\pi/2)$ remains. From (26), and taking into account that surge speed $u_c = 0$ for $\tilde{\alpha} = \pm(\pi/2)$, it is clear that in the observed case the derivative is always zero and the state remains unchanged. In other words, the trajectory is between the two regions of attraction in the phase plane. However, this is only the case in the approximated system, and the full system has a perturbation in the $\alpha$ degree of freedom, so it cannot stay identically at $\alpha = \pm(\pi/2)$. Taking all this into account, we conclude that the state $\tilde{\alpha}$ is locally asymptotically stable to a small neighborhood of the origin.

Next, the state $\tilde{r}_c$ is analyzed. Knowing that $\tilde{r}_c \geq 0$, from (33) it follows that $\dot{V}_{r_c} \leq 0$ for $\tilde{\alpha} \in (-(\pi/2), \pi/2)$. Since we have shown that state $\tilde{\alpha}$ converges to zero, yielding $u_c \neq 0$, $\tilde{r}_c$ cannot remain constantly at any $\tilde{r}_c > 0$ for $t \in (0, \infty)$.

Finally, we check the stability of the filter state $\tilde{x}_e$. The set for which the filter state should be attractive is defined with

$$\mathcal{E} := \left\{\tilde{x}_e \in \mathbb{R} : \tilde{x}_e = \frac{f(\tilde{r}_c, \tilde{\alpha})}{h}, \tilde{r}_c > 0, \tilde{\alpha} \in \left(-\frac{\pi}{2}, \frac{\pi}{2}\right)\right\}.$$

We observe, as in [27], that the state $\tilde{x}_e$ is linear and its origin is exponentially stable for $f(\tilde{r}_c, \tilde{\alpha}) = 0$. Therefore, if $f(\tilde{r}_c, \tilde{\alpha})$ is bounded then the solution $\tilde{x}_e(t)$ exists and is bounded with a gain $1/h$. Since we have shown that $\tilde{\alpha}$ tends to zero, consequently $f(\tilde{r}_c, \tilde{\alpha})$ also tends to zero.

Since the Lie bracket system (25)–(27) is locally uniformly asymptotically stable we conclude that the full system (24) is locally practically uniformly asymptotically stable to a small neighborhood of the origin.

## IV. Algorithm Implementation

In Section III, the stability of the proposed control algorithm is shown, however in practice there are additional limitations that have to be considered. As already noted in Section II-C, in a practical algorithm implementation, both range $r_c$ and relative angle $\alpha$ used in the surge speed control law for determining the $u_{\text{dir}}$ and $u_\zeta$ terms are not available since the source position is unknown. To deploy the algorithm in real-life situations, it is necessary to approximate the required information from the available measurements. Furthermore, the presence of a constant current acting on the system and modeling errors in the vehicle dynamics can cause the system to drift. This is a major issue, especially when the receivers are placed on the separate vehicles, which creates the need for an additional position control loop. These implementation issues are covered in Sections IV-A and IV-B, before the complete implementation of the TDOA source seeking scheme is presented.

### A. Approximation of Range $r_c$ and Relative Angle $\alpha$

To approximate information about the range $r_c$ and the relative angle $\alpha$ from the available measurements—in this case, the yaw rate $\dot{\psi}$ and the TDOA $\Delta$ measurements, the following second-order filter is proposed

$$\dot{v}_1 = -\omega_1 v_1 + \Delta \tag{36}$$

$$\dot{v}_2 = -\omega_2 v_2 + k_1\dot{\psi}(\Delta - \omega_1 v_1) \tag{37}$$

where $\omega_1$, $\omega_2$, and $k_1$ are tunable filter parameters. In the proposed filter, the state $v_1$ (36) is used to calculate the derivative of $\Delta$, whereas the additional lowpass filter (37) has a twofold purpose. It averages the signal to filter out the perturbations, but also to remove the brief changes in the signal sign caused by a small lag introduced by the derivative and the system dynamics that are present in a real-life scenario. The main idea behind the introduced filter is to extract the signal envelope value of the product $\dot{\psi}\dot{\Delta}$. Due to the introduced perturbation, the heading changes, and consequently $\Delta$ changes periodically during the algorithm operation. As the vehicle approaches the source, for the same change in the yaw degree of freedom a smaller change of $\Delta$ is achieved, until the system is finally above the source where there is no change in $\Delta$ for any change in $\psi$. This behavior is exploited to extract the information that is correlated with range $r_c$. Furthermore, to determine the surge speed sign, we observe the phase shift of the yaw rate and the time derivative of the $\Delta$ signal. When the source is in front of the baseline, both signals are in phase and their product is positive. When the source is behind, the signals are shifted by $\pi$ radians, and the resulting product is negative.

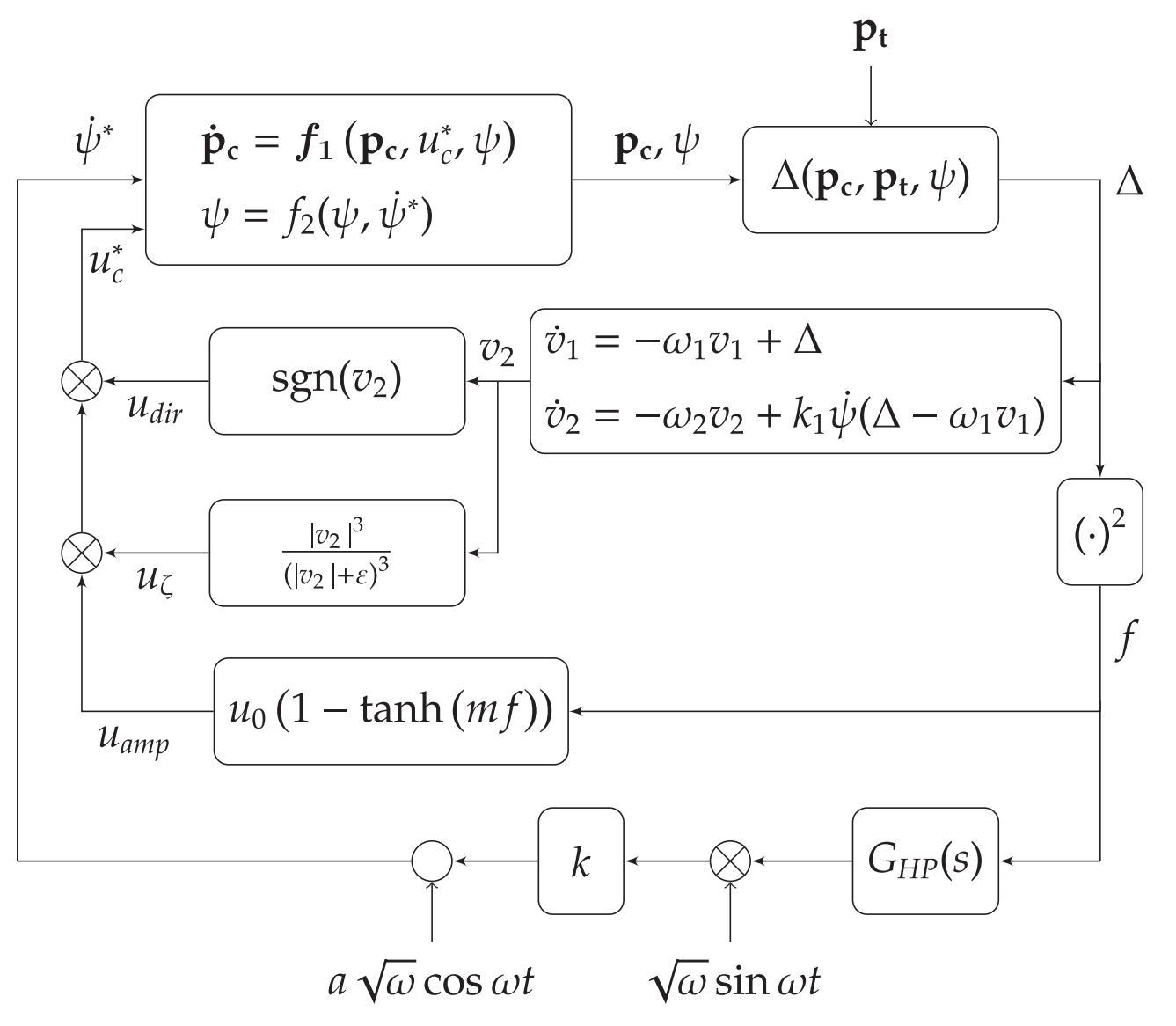

Fig. 2. TDOA source seeking scheme.

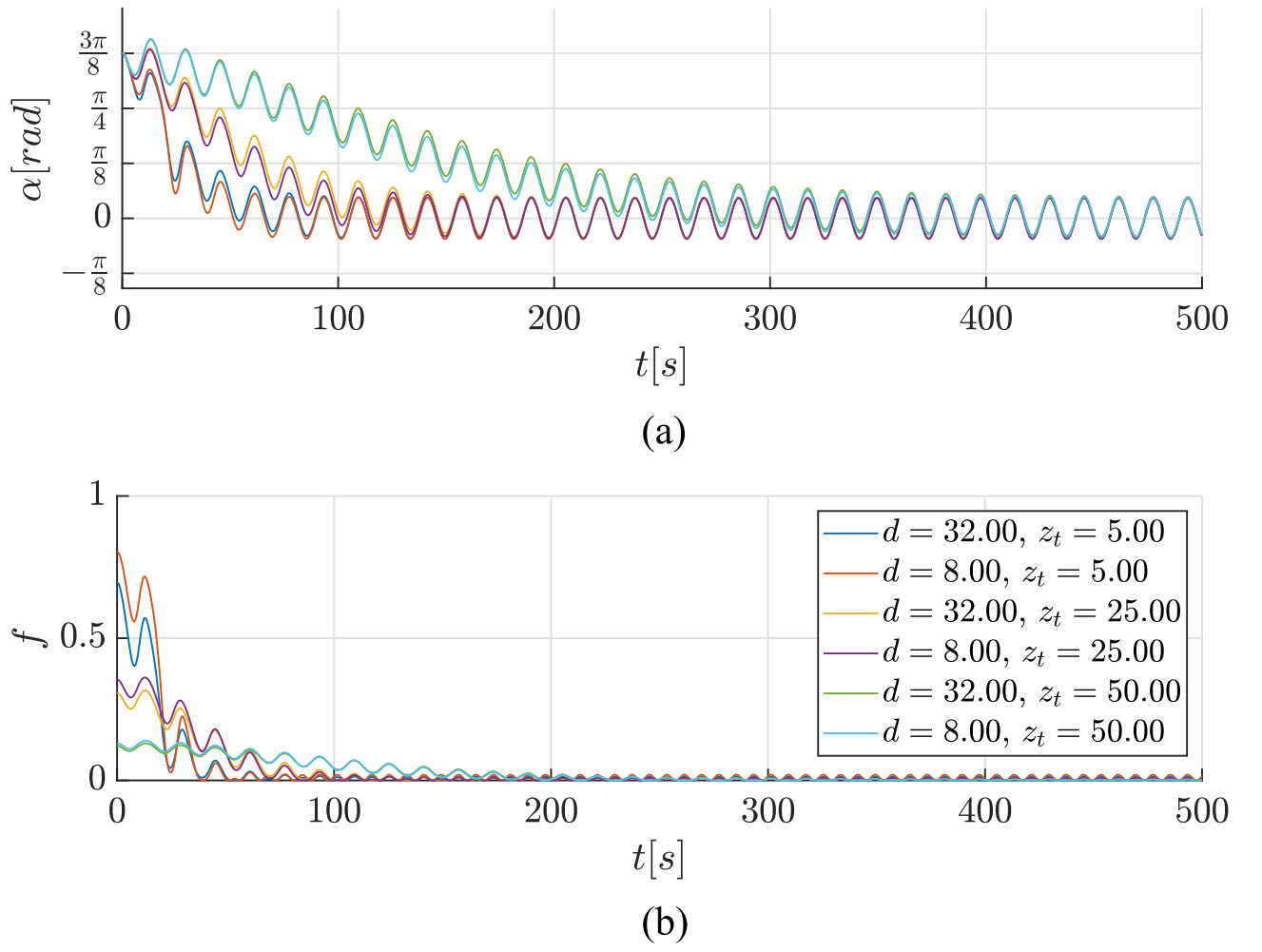

Fig. 3. Angle $\alpha$ and the cost $f$ response for the different values of depth $z_t$ and baseline $d$ with disabled surge speed controller and the source located at $\mathbf{p_t} = [0\ 0\ z_t]^T$. (a) Relative angle $\alpha$. (b) Cost value $f$.

Using the calculated value $v_2$, the respective terms in (18) are replaced with

$$u_{\mathrm{dir}} = \mathrm{sgn}\,(v_2),\ \text{and}\ u_\zeta = \frac{|v_2|^3}{(|v_2| + \varepsilon)^3}. \tag{38}$$

### B. Position Reference Control

The presence of a constant current acting on the system causes a constant position offset in the stationary state and a position drift while converging toward the source. To remove the static offset and improve the convergence speed when dealing with a constant current or other unmodeled disturbances, a position control loop is added, where the system kinematic model described with $\dot{x}_c^* = u_c^* \cos\psi$ and $\dot{y}_c^* = u_c^* \sin\psi$, is used to calculate the respective position references $x^*$ and $y^*$ from the velocity reference input $u_c^*$ and current baseline heading $\psi$. The vehicle's conventional control algorithms are then used to achieve the desired position. The addition of the position control loop introduces an additional delay inside the whole control loop, however that delay is negligible compared to the source seeking dynamics.

The complete TDOA source seeking control scheme is given in Fig. 2. In there, functions $\boldsymbol{f_1}\left(\mathbf{p_c}, u_c^*, \psi\right)$ and $f_2(\psi, \dot{\psi}^*)$ encapsulate the complete vehicle dynamics, including the low-level velocity controllers and position controllers if they are utilized. Notice that all the references for the low-level velocity and position controllers are marked with superscript "$*$." The transfer function $G_{\mathrm{HP}}(s)$ represents the highpass filter in the extremum seeking scheme.

## V. Simulation Results

Numerical simulations with the vehicle dynamics included were conducted to examine the behavior of the algorithm implementation depicted in Fig. 2. For the analysis purposes, the results of the yaw rate controller simulation with the surge speed controller turned off are first shown and discussed, before showing the results of the simulation of the full source seeking algorithm implementation.

### A. Yaw Rate Controller

Fig. 3 shows the response of the angle $\alpha$ and the cost value $f$ for the different source depths $z_t$ where only the yaw rate controller is active with parameters $a = 0.1\sqrt{\mathrm{rad/s}}$, $\omega = 2\pi/16\,\mathrm{rad/s}$, $k = -1.0\sqrt{\mathrm{rad/s}}$, and $h = 0.19\,1/\mathrm{s}$. As already discussed, for the same set of algorithm parameters, convergence is slower for the larger source depth. For the same perturbation amplitude, there is a smaller response in $\Delta$ and thus the extremum seeking controller takes more time to steer the system in the right direction. Due to the normalization of the TDOA measurement $\Delta_{\mathrm{TOA}}$ with the value of the baseline $d$, increasing the baseline four times has practically no influence on the convergence rate. However, it is a fraction slower for the larger baseline, as suggested in Section III. This is true for the ideal case without the measurement noise. A noise-influenced $\Delta$ measurement can be written as

$$\Delta = \frac{\Delta_{\mathrm{TOA}} v + \zeta_\Delta}{d} \tag{39}$$

where $\zeta_\Delta \sim \mathcal{N}(0, \sigma_\Delta^2)$ and the standard deviation $\sigma_\Delta$ are given in meters. By increasing the baseline $d$, the noise influence in the measurement is diminished, which justifies the idea of an adjustable baseline. It is important to note that increasing the baseline can introduce an additional noise, albeit small, to the measurement, due to the longer path that the acoustic signal needs to travel and which is not modeled in the presented case.

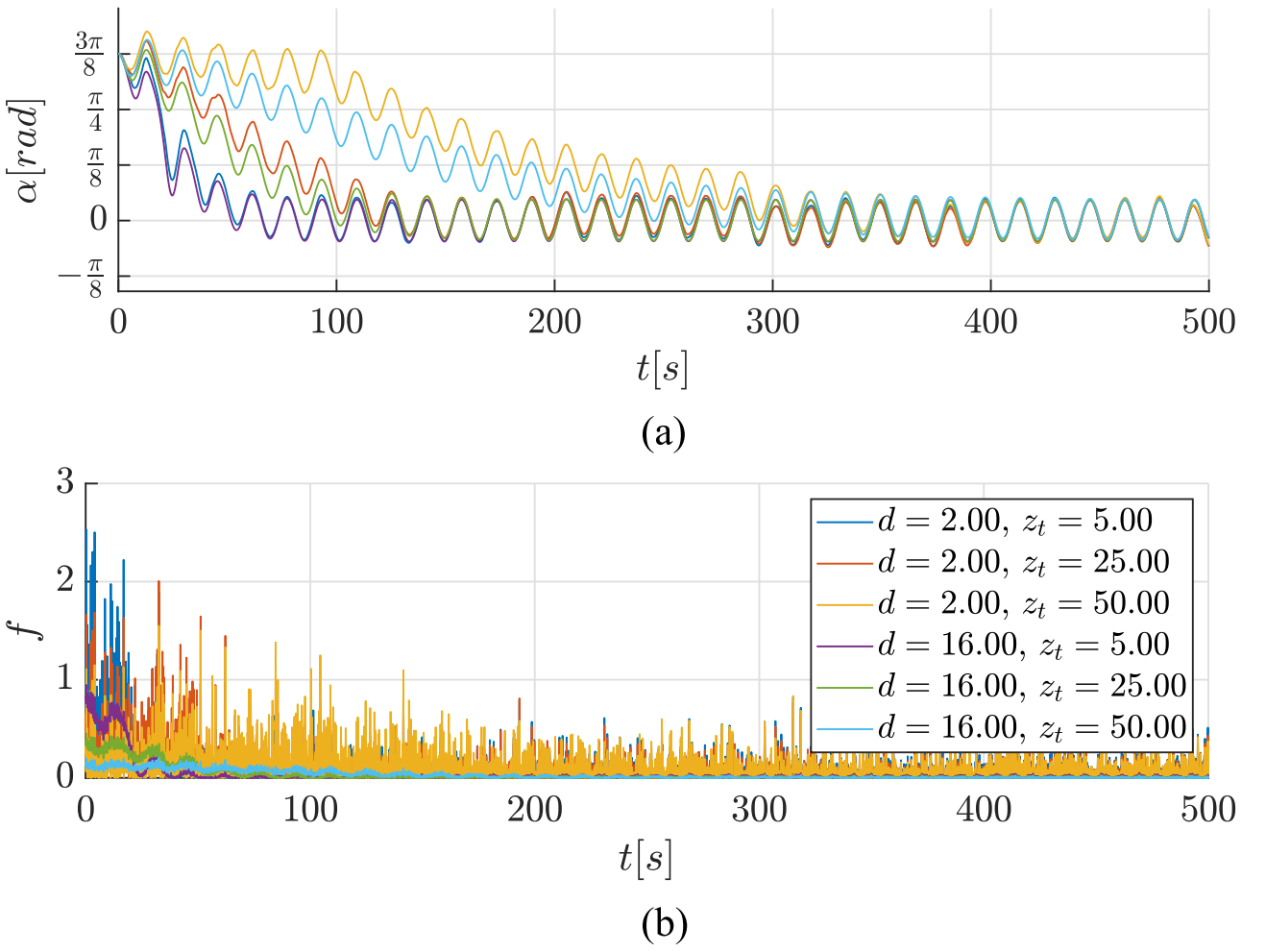

Fig. 4. Angle $\alpha$ and the cost $f$ response for the different values of depth $z_t$ and baseline $d$ with a measurement noise $\sigma_\Delta = 0.3$ m. The surge speed controller is disabled and the source is located at $\mathbf{p_t} = [0\ 0\ z_t]^T$. (a) Relative angle $\alpha$. (b) Cost value $f$.

In Fig. 4, the angle $\alpha$ and the cost $f$ responses for different values of depth $z_t$ and baseline $d$ with measurement noise $\sigma_\Delta = 0.3$ m are shown. The remaining parameters are the same as in Fig. 3. It is clearly visible that increasing the system baseline reduces the influence of the measurement noise present in the $\Delta$ measurement. The angle $\alpha$ convergence is faster for the larger baseline, which is especially notable for the larger source depths.

## B. Source Seeking Algorithm

To test the full algorithm behavior with the scheme introduced in Section IV, a set of simulation experiments was done. Three cases are observed in detail: the system behavior with and without the measurement noise present in the $\Delta$ measurement, and finally, the system behavior with a constant current acting on the system.

*1) No Measurement Noise:* In Fig. 5, the trajectory and range response of the system for different values of depth $z_t$ and baseline $d$ is shown with parameters $a = 0.15\sqrt{\mathrm{rad/s}}$, $\omega = 2\pi/16\,\mathrm{rad/s}$, $k = -1.0\sqrt{\mathrm{rad/s}}$, $h = 0.19$ 1/s, $u_0 = 0.5$ m/s, $m = 100$, $\varepsilon = 4$ m, and $q = 3$, whereas the second-order filter parameters are $\omega_1 = 0.8$ 1/s, $\omega_2 = 0.15$ 1/s, and $k_1 = 1000$. As already discussed in Section III, increasing the baseline $d$ does not significantly influence the convergence speed of the system. From Fig. 5(b), we observe that initially the range $r_c$ is constant, while the baseline slowly turns toward the target. When the angle $\alpha$ is small enough, as defined by the parameter $m$, the surge speed increases and the system starts approaching the source. The baseline surge speed $u_c$ is shown in Fig 6(c). In Fig. 6(b), it can be seen that while the angle $\alpha$, and consequently the cost $f$, is large, there is no surge speed $u_c$. It can also be noted that the surge speed keeps oscillating as the heading changes due to perturbations, but its average value decreases as the baseline approaches the signal source, i.e., it tends toward zero as the range to the source approaches zero, showing that the introduced second-order filter successfully approximates the distance information.

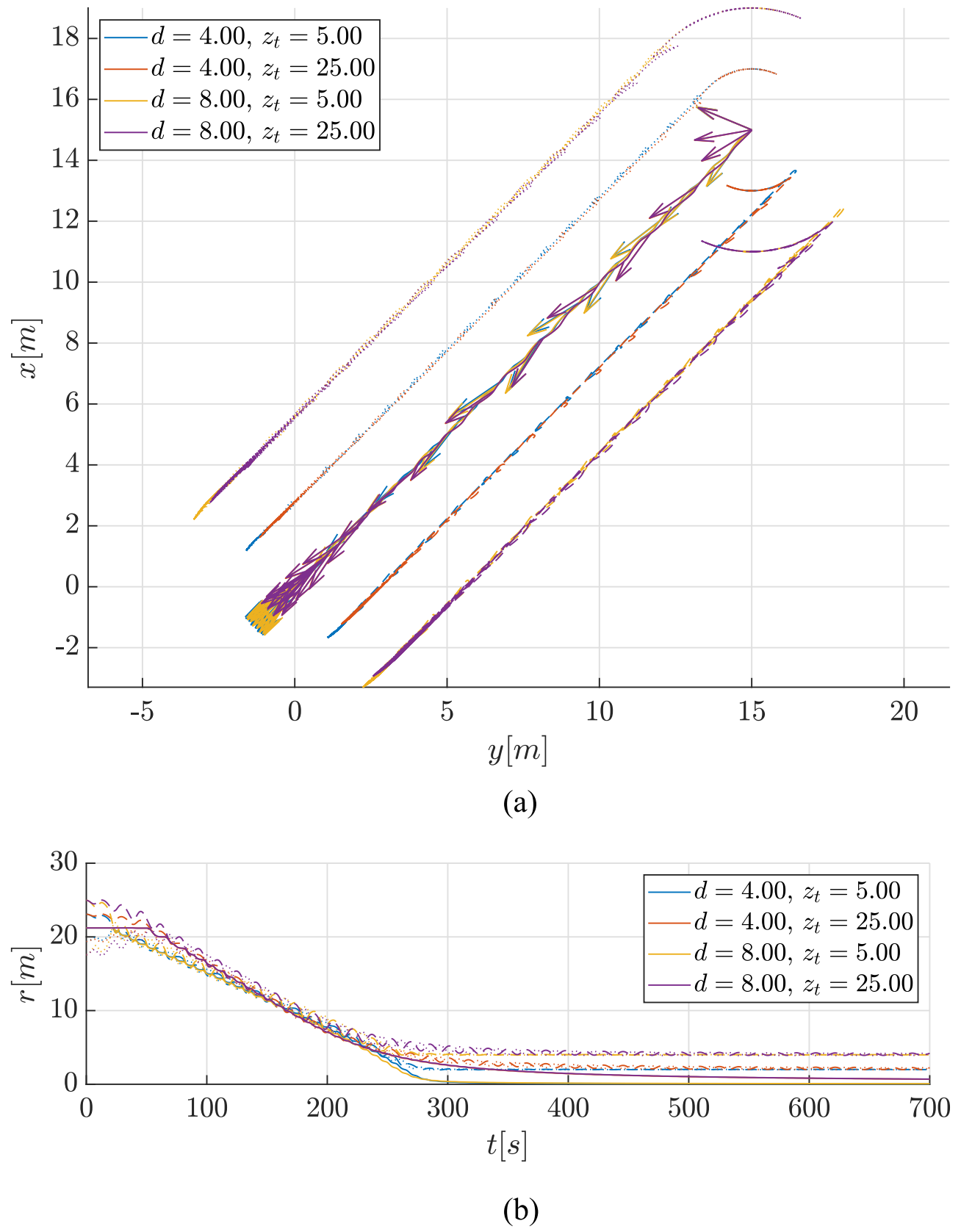

Fig. 5. Trajectory and range responses for the different values of source depth $z_t$ and baseline $d$ with the source located at $\mathbf{p_t} = [0\ 0\ z_t]^T$. (a) Acoustic receiver $\mathbf{p_1}$ trajectory (dashed), acoustic receiver $\mathbf{p_2}$ trajectory (dotted), baseline center $\mathbf{p_c}$ trajectory (solid). (b) Acoustic receiver $\mathbf{p_1}$ range (dashed), acoustic receiver $\mathbf{p_2}$ range (dotted), baseline center $\mathbf{p_c}$ range (solid).

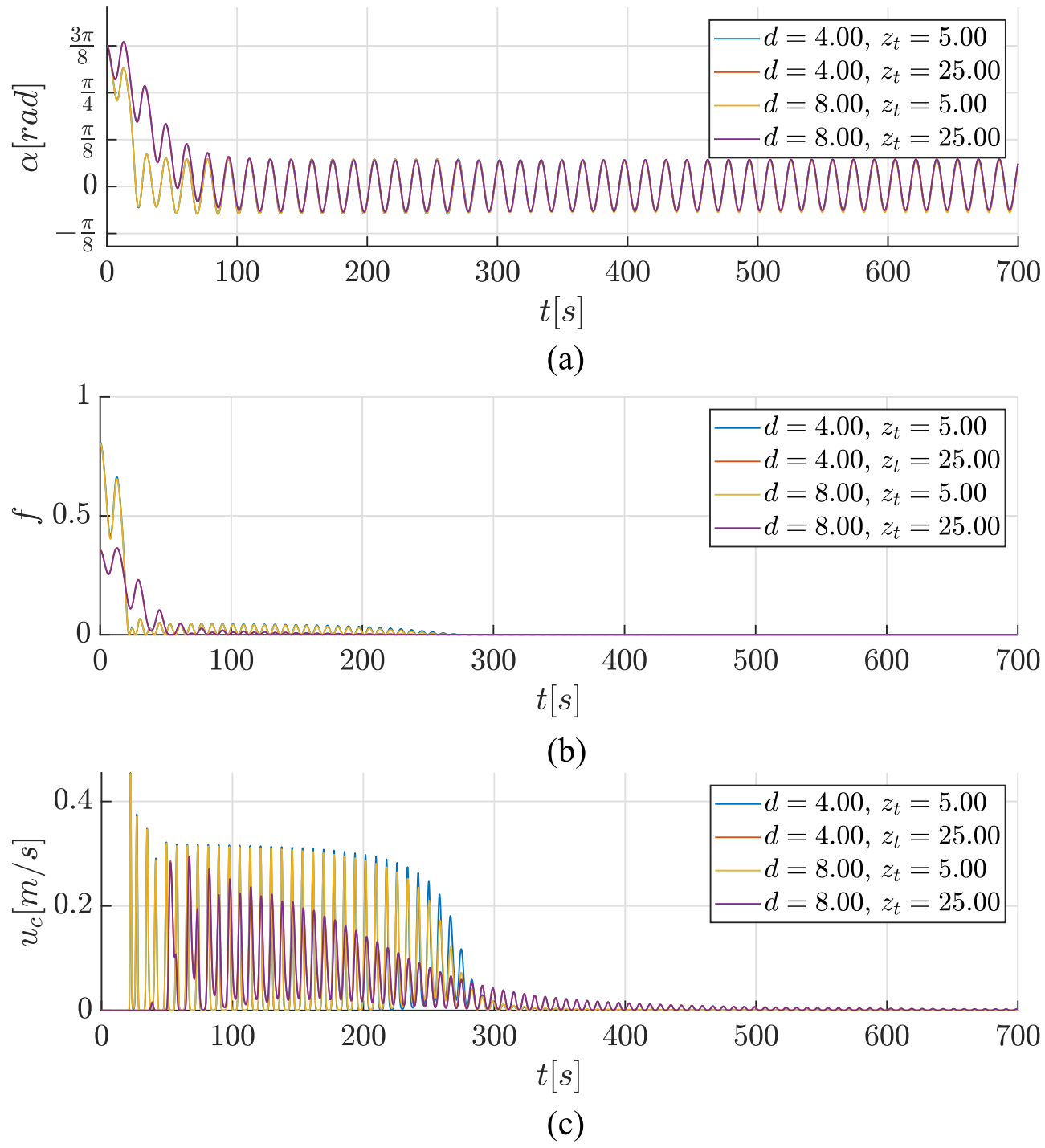

Fig. 6. (a) Relative angle $\alpha$, (b) cost value $f$, and (c) surge speed $u_c$ response for the different values of source depth $z_t$ and baseline $d$ with the source located at $\mathbf{p_t} = [0\ 0\ z_t]^T$.

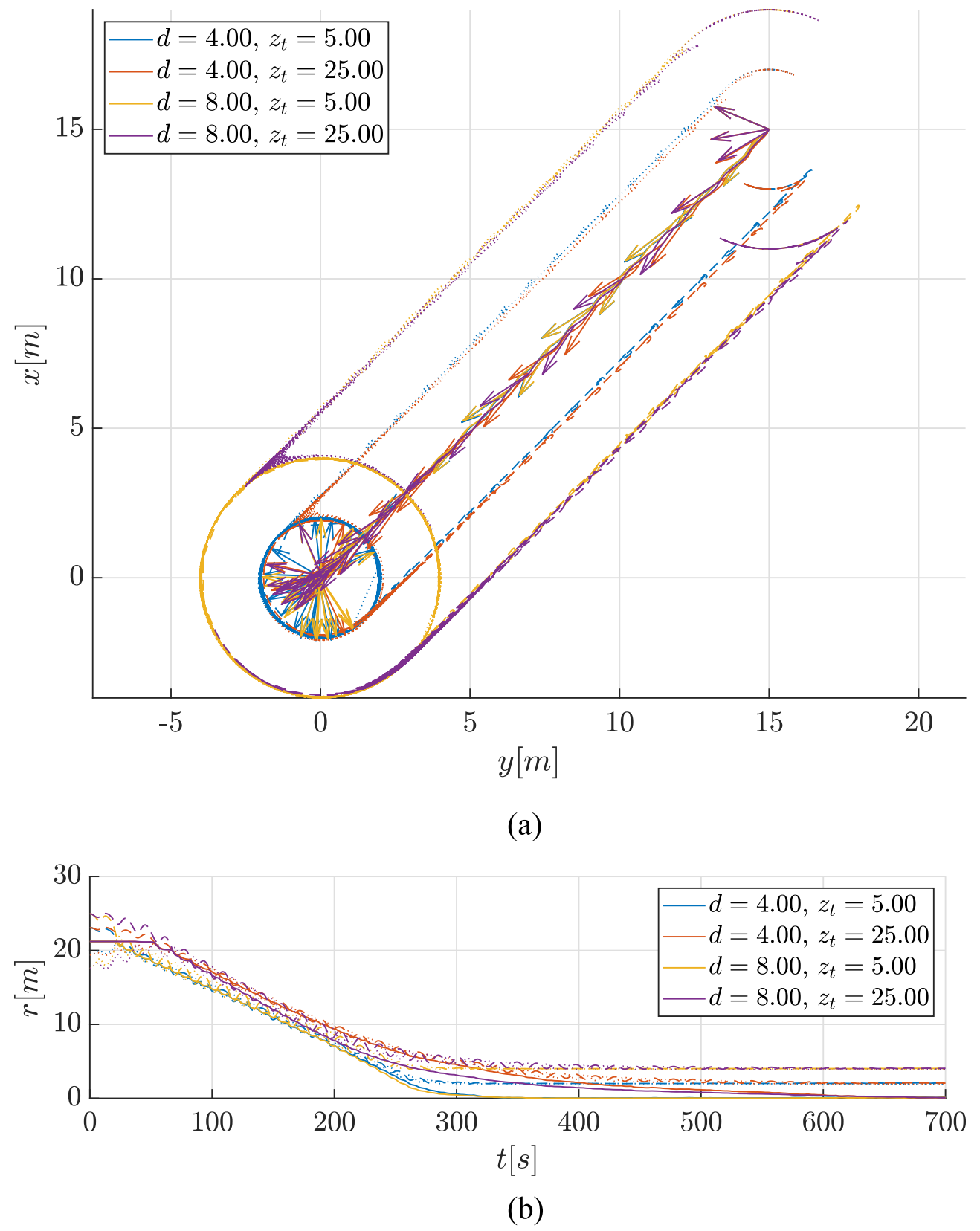

Fig. 7. Trajectories and ranges response for different values of the source depth $z_t$ and the baseline $d$ with measurement noise $\zeta_\Delta = 0.3$ m and with the source located at $\mathbf{p_t} = [0\ 0\ z_t]^T$. (a) Acoustic receiver $\mathbf{p_1}$ trajectory (dashed), acoustic receiver $\mathbf{p_2}$ trajectory (dotted), baseline center $\mathbf{p_c}$ trajectory (solid). (b) Acoustic receiver $\mathbf{p_1}$ range (dashed), acoustic receiver $\mathbf{p_2}$ range (dotted), baseline center $\mathbf{p_c}$ range (solid).

*2) With Measurement Noise:* In Fig. 7, the trajectory and range response of the system for different values of the target depth $z_t$ and the baseline $d$ are shown with measurement noise $\sigma_\Delta = 0.3$ m included, while all the other parameters are the same as in the noiseless case. As expected, convergence is faster for the larger baseline. The benefit of a larger baseline is especially evident when looking at the surge speed direction in Fig. 8(c). Namely, there are many incorrect surge speed direction changes caused by the noisy $\Delta$ measurement in the case of the smaller baseline. In Fig. 4(a), it is noticeable that when the system is very close to the source, angle $\alpha$ starts to change rapidly. However, this is a numerical instability that happens due to the choice of the coordinate system since $\alpha$ is not defined for $r_c = 0$. Practically, this effect has no consequences on algorithm operation, but increasing the parameter $m$ can reduce it. For larger values of the parameter $m$ the baseline center deviates less from the line connecting the initial baseline position and the source position, but it also takes more time for the system to converge to the source's position due to a lower average speed.

*3) Constant Current:* In Fig. 9, the trajectory and range responses of the system with the velocity reference control and the system with the position reference control are compared for the baseline $d = 5\,\mathrm{m}$, with a constant current $v_c = [0.05\ \ 0\ \ 0]^T\,\mathrm{m/s}$ acting on the vehicles. It is visible that the system with the position reference control has nearly the same performance as in the case without the currents, while the velocity reference controlled system cannot compensate for the disturbance. Even though the disturbance is small, the baseline slowly drifts and it takes twice as much time for the system to arrive near the signal source, with a constant offset present when it finally does reach it, as seen in the range response in Fig. 9(b). The surge speed response in Fig. 10(c) shows that the velocity reference controlled system, when close to the source, achieves the higher amplitudes of the surge speed necessary to counter the influence of the constant current.

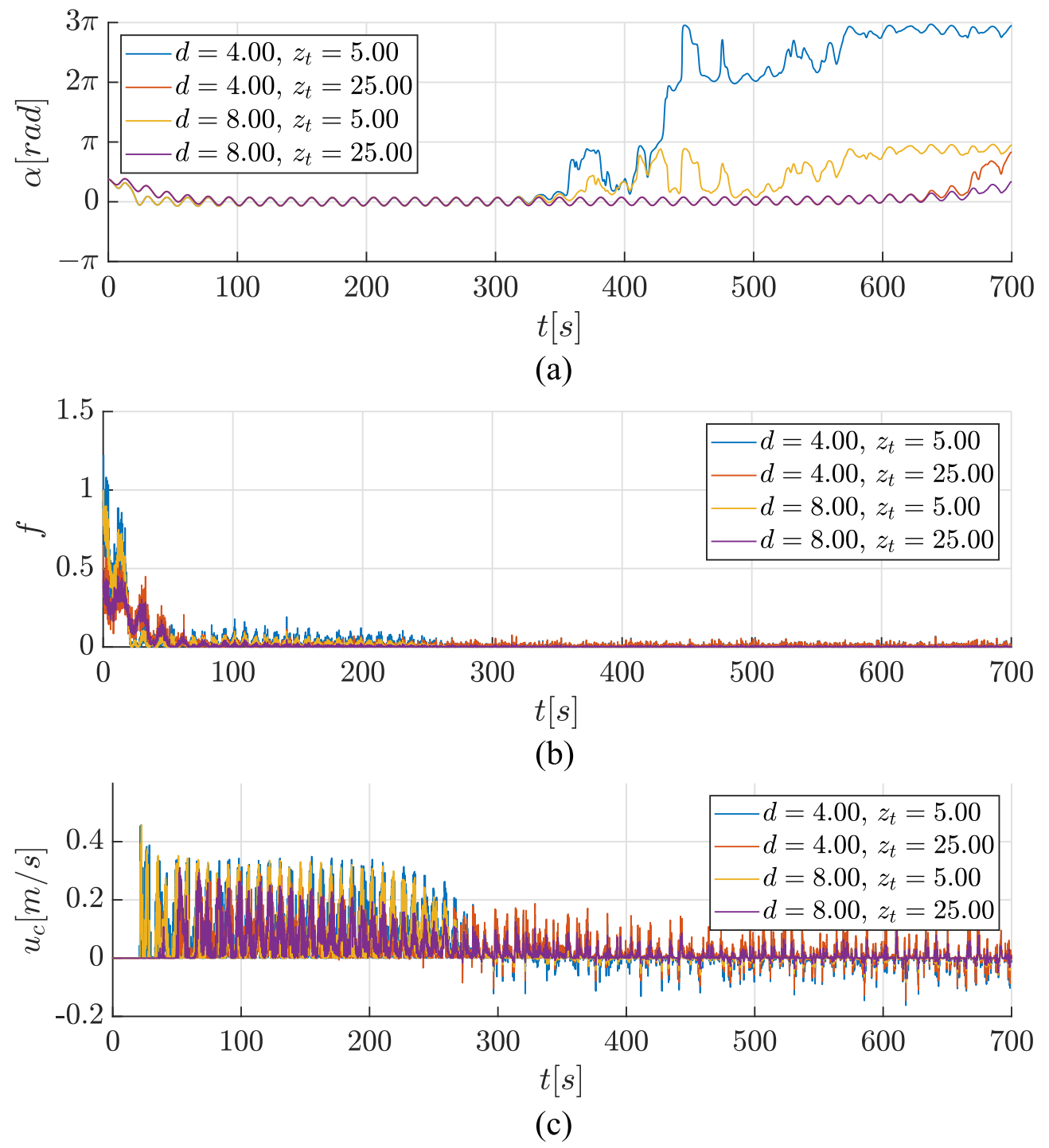

Fig. 8. (a) Relative angle $\alpha$, (b) cost value $f$, and (c) surge speed $u_c$ response for the different values of source depth $z_t$ and baseline $d$ with measurement noise $\zeta_\Delta = 0.3$ m and with the source located at $\mathbf{p_t} = [0\ 0\ z_t]^T$.

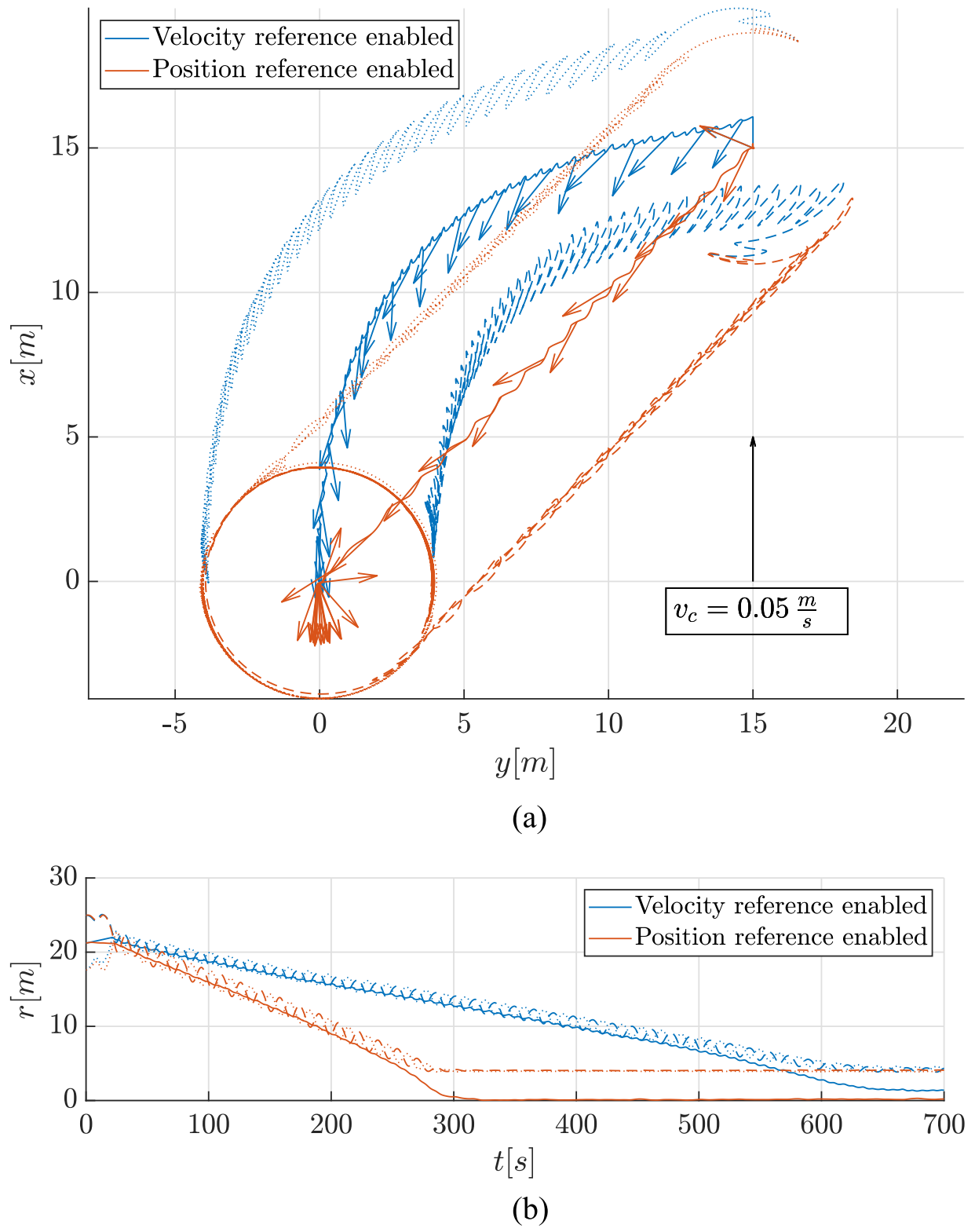

Fig. 9. Trajectories and ranges comparison for the position reference and the velocity reference control with a constant current $v_c = 0.05$ m/s and with the source located at $\mathbf{p_t} = [0\ 0\ 5]^T$. (a) Acoustic receiver $\mathbf{p_1}$ trajectory (dashed), acoustic receiver $\mathbf{p_2}$ trajectory (dotted), baseline center $\mathbf{p_c}$ trajectory (solid). (b) Acoustic receiver $\mathbf{p_1}$ range (dashed), acoustic receiver $\mathbf{p_2}$ range (dotted), baseline center $\mathbf{p_c}$ range (solid).

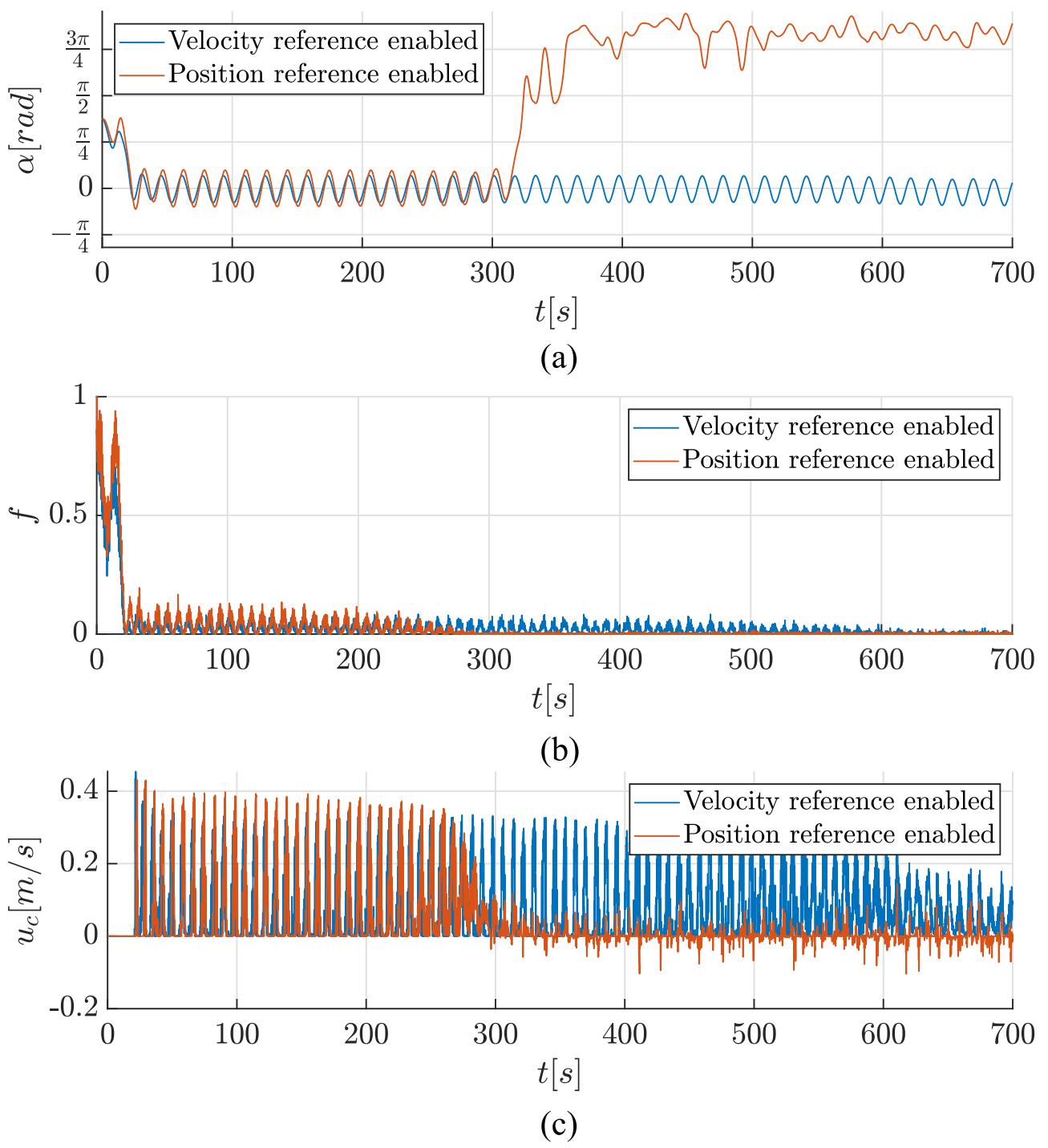

Fig. 10. (a) Relative angle $\alpha$, (b) cost value $f$, and (c) surge speed $u_c$ response for the different values of source depth $z_t$ and baseline $d$ with a constant current $v_c = 0.05\,\mathrm{m/s}$ and with the source located at $\mathbf{p_t} = [0\ 0\ 5]^T$.

## VI. Experimental Results

A small set of proof-of-concept experiments was conducted in November 2017 at Jarun lake in Zagreb, Croatia. The experimental setup consisted of two autonomous over-actuated marine surface platforms *aPad*, shown in Fig. 11 and developed in the Laboratory for Underwater Systems and Technologies [28]. The *aPad* vehicle is a small scale over-actuated unmanned surface marine vehicle capable of an omnidirectional motion. It is equipped with four thrusters in an "X" configuration that enables the motion in the horizontal plane under any orientation. The vehicle is 0.385 m high, 0.756 m wide, and long, and weighs approximately 25 kg. Both platforms were identically equipped with a Nanomodem, a low-cost acoustic modem developed at Newcastle University [29]. An additional Nanomodem was used as the acoustic source emitter, and it was positioned on the lakebed at a depth of approximately 3 m.

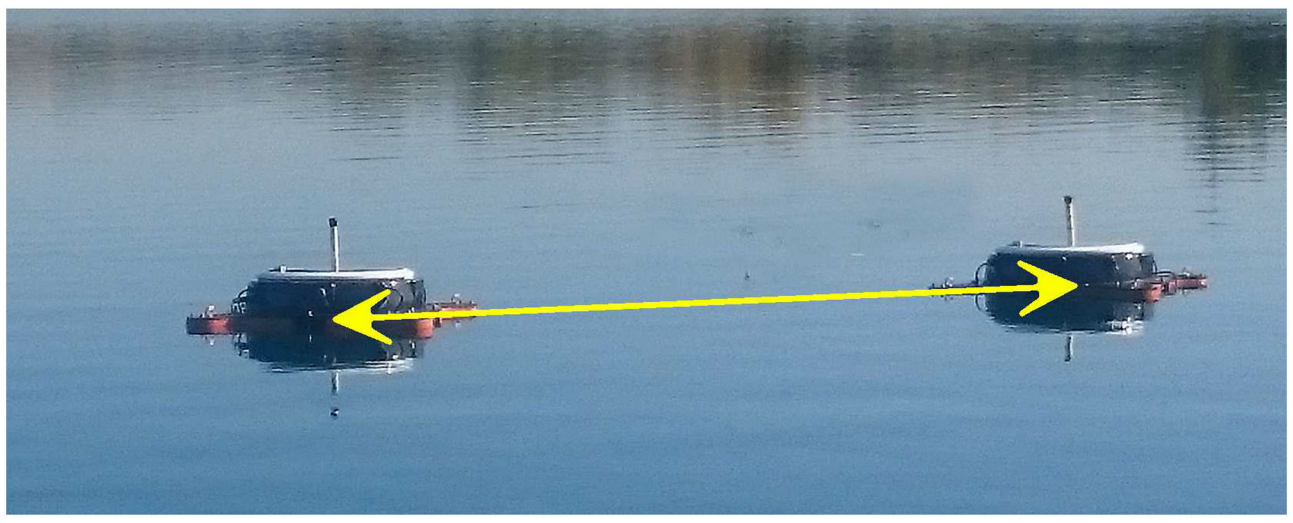

Fig. 11. Two autonomous over-actuated marine surface platforms *aPad* equipped with acoustic receivers performing the experiment. The receiver baseline is marked with a yellow arrow.

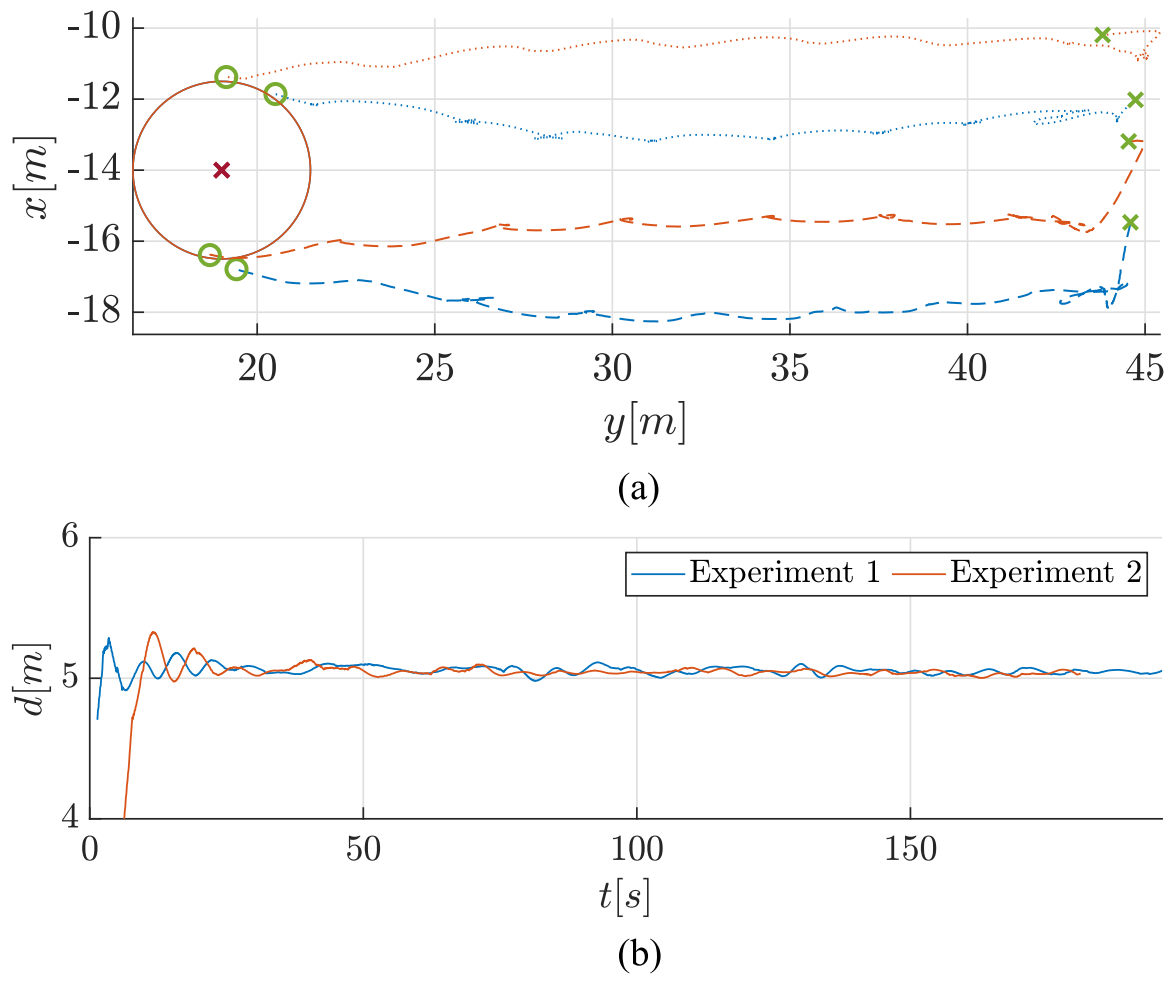

Fig. 12. Trajectory and baseline responses of the real-life source seeking experiments with the baseline reference $d = 5\,\mathrm{m}$ and the source located at $\mathbf{p_t} \approx [-14\ 19\ 3]^T$. (a) Acoustic receiver $\mathbf{p_1}$ trajectory (dashed), and the acoustic receiver $\mathbf{p_2}$ trajectory (dotted). Green crosses denote the receivers' starting position, green circles denote ending position, while the red cross denotes source $\mathbf{p_t}$ surrounded with a 5 m diameter circle. (b) Baseline $d$.

A precise measurement of the acoustic signal TOA was acquired in the following way: every two seconds, the Nanomodem positioned on the lakebed broadcast an acoustic package that was received by the receivers on the vehicles. When the header of the acoustic packet was decoded on the receiver side, a digital pin on the receiver, directly connected to the time mark input of the u-blox NEO-M8T chip, was triggered and a precise GPS timestamp was recorded on the same chip. This configuration made a very low-latency measurements possible. The recorded timestamp was then exchanged between the two platforms using the Wi-Fi communication and used in the TDOA calculation.

The results of the two conducted experiments are shown in Figs. 12 and 13. During the experiments, the following parameters were used: $a = 0.05\,\sqrt{\mathrm{rad/s}}$, $\omega = 2\pi/30\,\mathrm{rad/s}$, $k = -0.125\,\sqrt{\mathrm{rad/s}}$, $h = 0.1\,1/\mathrm{s}$, $u_0 = 0.3\,\mathrm{m/s}$, $d = 5\,\mathrm{m}$, $m = 20$, $\varepsilon = 0.5\,\mathrm{m}$, and $q = 3$, whereas the second-order filter parameters were $\omega_1 = 0.8\,1/\mathrm{s}$, $\omega_2 = 0.15\,1/\mathrm{s}$, and $k_1 = 1000$. These experiments used the position reference control to compensate for the outside disturbances and model errors. In Fig. 12(b), the formation baseline $d$ response is shown and it is clearly visible that after an initial transient, position controllers employed

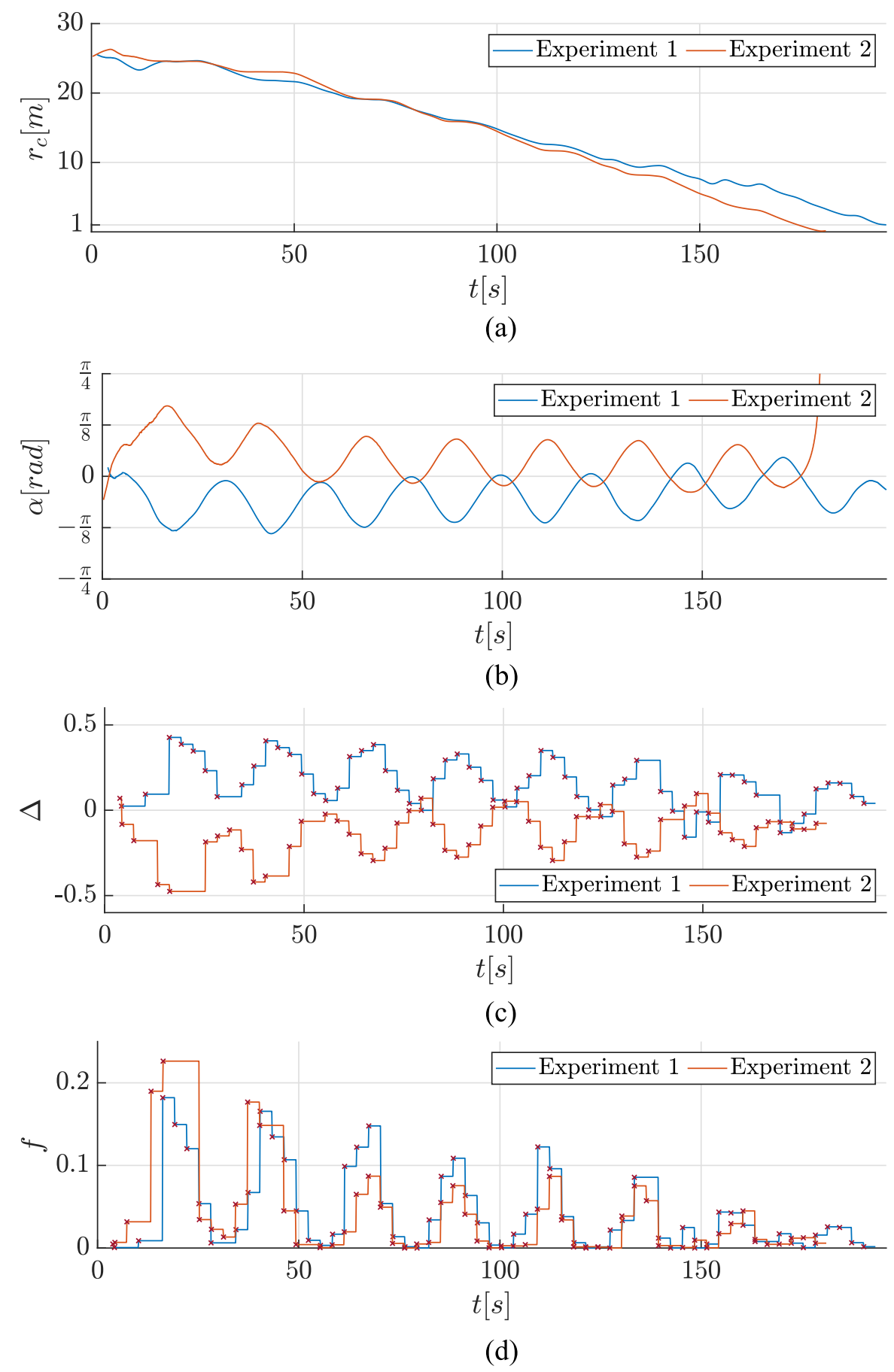

Fig. 13. Range $r_c$, relative angle $\alpha$, $\Delta$ measurement, and cost function responses of the real-life source seeking experiments with the baseline reference $d = 5$ m and the source located at $\mathbf{p_t} \approx [-14\ 19\ 3]^T$. (a) Range $r_c$. (b) Relative angle $\alpha$. (c) TDOA $\Delta$. (d) Cost function value $f$.

on the respective vehicles keep the two receivers at the desired distance of 5 m.

During the experiments, TDOA measurements were rejected when the measurement was larger than the current baseline $d$ because then it could be safely assumed that a spurious measurement due to the signal reflection or some other interference had arrived. Looking at the normalized $\Delta$ measurement, shown in Fig 13(c), it is clear that acquired measurements are intermittent. In spite of this, the formation successfully converged toward the acoustic source, as can be seen in Fig 13(a). This was achieved by selecting a sufficiently large perturbation frequency for the extremum seeking controller.

## VII. Conclusion

The proposed algorithm can be used in cases where an underwater acoustic signal source needs to be approached and localized in the horizontal plane, and only two acoustic receivers are available, with no additional means of acquiring range measurements that could otherwise be used for the localization. As shown herein, the initial experimental results suggest that the proposed control scheme can be used in a real-life environment. The authors have found that the algorithm parameters are intuitive and easy to tune to achieve robust behavior in the presence of the measurement noise. Also, despite the intermittent and delayed nature of the acoustic signals, by selecting a sufficiently large perturbation period, these effects can be successfully mitigated. Although convergence speed of the baseline orientation is currently satisfactory, considering the algorithm is intended for a marine environment, it could be further improved by using a more advanced extremum seeking schemes in the yaw degree of freedom. Further work will include comprehensive experimental tests where the influence of different variables on the algorithm performance will be inspected, such as setting the different target depths, or using the different baselines.

## Appendix

In [27], a couple of assumptions on the vector fields $\boldsymbol{b_i}$ are given. The first one is that $\boldsymbol{b_i} \in C^2 : \mathbb{R} \times \mathbb{R}^n \to \mathbb{R}^n$ for $i = 0, \ldots, m$. In the case of the system (24) terms $\boldsymbol{b_0}$, $\boldsymbol{b_1}$, and $\boldsymbol{b_2}$ are sufficiently smooth vector fields if a simple modification is done. Namely, the signum function in (18) is discontinuous but for the mathematical analysis it can be easily approximated with a continuous function, i.e., $\mathrm{sgn}(x) \approx \tanh(\mu x)$ if parameter $\mu \gg 1$.

The second assumption is that for every compact set $\mathcal{C} \subseteq \mathbb{R}^n$ there exist $\boldsymbol{A_1}, \ldots, \boldsymbol{A_6} \in [\boldsymbol{0}, \infty)$ such that

$$\left|\boldsymbol{b_i}(t,\boldsymbol{\eta})\right| \le \boldsymbol{A_1}, \left|\frac{\partial \boldsymbol{b_i}(t,\boldsymbol{\eta})}{\partial t}\right| \le \boldsymbol{A_2}, \left|\frac{\partial \boldsymbol{b_i}(t,\boldsymbol{\eta})}{\partial \boldsymbol{\eta}}\right| \le \boldsymbol{A_3}$$
$$\left|\frac{\partial^2 \boldsymbol{b_j}(t,\boldsymbol{\eta})}{\partial t \partial \boldsymbol{\eta}}\right| \le \boldsymbol{A_4}, \left|\frac{\partial [\boldsymbol{b_j},\boldsymbol{b_k}](t,\boldsymbol{\eta})}{\partial \boldsymbol{\eta}}\right| \le \boldsymbol{A_5}$$
$$\left|\frac{\partial [\boldsymbol{b_j},\boldsymbol{b_k}](t,\boldsymbol{\eta})}{\partial t}\right| \le \boldsymbol{A_6},$$
$$\boldsymbol{\eta} \in \mathcal{C}, t \in \mathbb{R}, i = 0, \ldots, m, j = 1, \ldots, m, k = j, \ldots, m.$$

For the system (21) state vector is defined with $\boldsymbol{\eta} = [r_c\ \ \alpha\ \ x_e]^T$. Going further, without loss of generality, it is assumed that $q = 1$ and $d = 1$. Given inequalities are tested for vectors $\boldsymbol{b_0}$, $\boldsymbol{b_1}$, and $\boldsymbol{b_2}$, starting with $|\boldsymbol{b_0}(t,\boldsymbol{\eta})|$

$$\left|\left[-\frac{r_c u_{\mathrm{dir}} u_{\mathrm{amp}}}{r_c+\varepsilon}\cos\alpha \quad \frac{u_{\mathrm{dir}} u_{\mathrm{amp}}}{r_c+\varepsilon}\sin\alpha \quad -x_e h + f\right]^T\right| \le \boldsymbol{A_1}. \tag{40}$$

The variables $u_{\mathrm{dir}} \in [-1, 1]$ and $u_{\mathrm{amp}} \in [0, u_0)$, and functions $|\sin\alpha| \le 1$ and $|\cos\alpha| \le 1$ are bounded for all $\boldsymbol{\eta} \in \mathcal{C}$, while the parameter $h$ is a constant. Since $\varepsilon > 0$ and $r_c \ge 0$, denominator $(r_c + \varepsilon)$ is larger than zero and it is clear that bound $\boldsymbol{A_1}$ exists. The solution for inequality $|b_2(t,\boldsymbol{\eta})| = |[0\ \ a\ \ 0]^T| \le \boldsymbol{A_1}$ is trivial, while in the inequality $|\boldsymbol{b_1}(t,\boldsymbol{\eta})| = |[0\ \ k(f - x_e h)\ \ 0]^T| \le \boldsymbol{A_1}$, variable $f \in [0, 1]$ and for the $x_e \in \mathcal{C}$ bound also exists. Next, boundedness of

$$\left|\frac{\partial \boldsymbol{b_0}(t,\boldsymbol{\eta})}{\partial \boldsymbol{\eta}}\right| = \left|\left[\frac{\partial \boldsymbol{b_0}(t,\boldsymbol{\eta})}{\partial r_c} \quad \frac{\partial \boldsymbol{b_0}(t,\boldsymbol{\eta})}{\partial \alpha} \quad \frac{\partial \boldsymbol{b_0}(t,\boldsymbol{\eta})}{\partial x_e}\right]^T\right|$$

$$\left\| \begin{bmatrix} 0 & \frac{8akr_c\left(16\sin(2\alpha)r_c^2{z_t}^2+2\sin(4\alpha)r_c^2+16\sin(2\alpha){z_t}^4+8\sin(2\alpha){z_t}^2+\sin(2\alpha)\right)}{r_1^3r_2^3} & 0 \\ 0 & \frac{8akr_c^2\left(16r_c^4\cos(2\alpha)+32r_c^2{z_t}^2\cos(2\alpha)+4r_c^2\cos(2\alpha)^2+4r_c^2+16{z_t}^4\cos(2\alpha)+8{z_t}^2\cos(2\alpha)+\cos(2\alpha)\right)}{r_1^3r_2^3} & 0 \\ 0 & 0 & 0 \end{bmatrix}^T \right\| \leq \boldsymbol{A_5}. \tag{43}$$

is investigated. Calculating the partial derivative $(\partial \boldsymbol{b_0}(t, \boldsymbol{\eta}))/\partial r_c$ yields vector

$$\begin{bmatrix} -\frac{r_c\cos(\alpha)u_{\text{dir}}u_{\text{amp}}}{(r_c+\varepsilon)^2}+\frac{\cos(\alpha)u_{\text{dir}}u_{\text{amp}}}{r_c+\varepsilon}-\frac{r_c\sin(2\alpha)\cos(\alpha)u_{\text{amp}}\left({u_{\text{dir}}}^2-1\right)\sigma_1}{r_c+\varepsilon} \\ +\frac{2r_c\cos(\alpha)u_{\text{dir}}\left({\sigma_2}^2-1\right)\sigma_1\Delta}{r_c+\varepsilon} \\ \frac{\sin(\alpha)u_{\text{dir}}u_{\text{amp}}}{(r_c+\varepsilon)^2}-\frac{2\sin(\alpha)u_{\text{dir}}\left({\sigma_2}^2-1\right)\sigma_1\Delta}{r_c+\varepsilon}+\frac{\sin(2\alpha)\sin(\alpha)u_{\text{amp}}\left({u_{\text{dir}}}^2-1\right)\sigma_1}{r_c+\varepsilon} \\ 2\sigma_1\Delta \end{bmatrix} \tag{41}$$

where

$$\sigma_1 = \frac{2r_c+\sin(\alpha)}{2r_1} - \frac{2r_c-\sin(\alpha)}{2r_2}$$

and $\sigma_2 = \tanh(\Delta^2)$, while the partial derivative $(\partial \boldsymbol{b_0}(t, \boldsymbol{\eta}))/\partial \alpha$ is given with

$$\begin{bmatrix} -\frac{r_c\sin(\alpha)u_{\text{dir}}u_{\text{amp}}}{r_c+\varepsilon}-\frac{r_c\cos(\alpha)\sigma_1u_{\text{amp}}\left({u_{\text{dir}}}^2-1\right)}{r_c+\varepsilon}+\frac{2r_c\cos(\alpha)u_{\text{dir}}\left({\sigma_2}^2-1\right)\sigma_3\Delta}{r_c+\varepsilon} \\ -\frac{\cos(\alpha)u_{\text{dir}}u_{\text{amp}}}{r_c+\varepsilon}+\frac{\sin(\alpha)\sigma_1u_{\text{amp}}\left({u_{\text{dir}}}^2-1\right)}{r_c+\varepsilon}-\frac{2\sin(\alpha)u_{\text{dir}}\left({\sigma_2}^2-1\right)\sigma_3\Delta}{r_c+\varepsilon} \\ \frac{r_c^2\sin(2\alpha)}{r_1r_2} \end{bmatrix} \tag{42}$$

where

$$\sigma_1 = \sin(2\alpha)\sigma_3 + 2\cos(2\alpha)\Delta$$

$$\sigma_2 = \tanh(\Delta^2) \quad \text{and}$$

$$\sigma_3 = \frac{r_c\cos(\alpha)}{2r_1} + \frac{r_c\cos(\alpha)}{2r_2}$$

and the third partial derivative is

$$\frac{\partial \boldsymbol{b_0}(t, \boldsymbol{\eta})}{\partial x_e} = [0 \ \ 0 \ \ -h]^T.$$

For the calculated partial derivatives, the same bounds on terms $u_{\text{amp}}$, $u_{\text{dir}}$, $\varepsilon$, $h$ and $f$ apply as in (40). Additionally, variable $\Delta \in [-1, 1]$. Since $z_t > 0$, ranges $r_1$ and $r_2$ are greater than zero for every $\boldsymbol{\eta} \in \mathcal{C}$ and in the following inequality

$$\left| \frac{\partial \boldsymbol{b_0}(t, \boldsymbol{\eta})}{\partial \boldsymbol{\eta}} \right| \leq \boldsymbol{A_3}$$

bound $\boldsymbol{A_3}$ exists. Since $\boldsymbol{b_0}$, $\boldsymbol{b_1}$, and $\boldsymbol{b_2}$ do not explicitly depend on $t$, the conditions

$$\left| \frac{\partial \boldsymbol{b_i}(t, \boldsymbol{\eta})}{\partial t} \right| \leq \boldsymbol{A_2} \text{ and } \left| \frac{\partial^2 \boldsymbol{b_j}(t, \boldsymbol{\eta})}{\partial t \partial \boldsymbol{\eta}} \right| \leq \boldsymbol{A_4}$$

are satisfied trivially. The following inequality

$$\left| \frac{\partial [\boldsymbol{b_1}, \boldsymbol{b_2}](t, \boldsymbol{\eta})}{\partial \boldsymbol{\eta}} \right| \leq \boldsymbol{A_5}$$

where

$$[\boldsymbol{b_1}, \boldsymbol{b_2}] = \begin{bmatrix} 0 & ak\frac{r_c^2\sin(2\alpha)}{r_1r_2} & 0 \end{bmatrix}^T,$$

is shown in (43) at the top of this page, and the same argumentation is used as in (41), ranges $r_1$ and $r_2$ are greater than zero for every $\boldsymbol{\eta} \in \mathcal{C}$, and bound $\boldsymbol{A_5}$ exists. Finally

$$\left| \frac{\partial [\boldsymbol{b_1}, \boldsymbol{b_2}](t, \boldsymbol{\eta})}{\partial t} \right| \leq \boldsymbol{A_6}$$

is satisfied trivially since $[\boldsymbol{b_1}, \boldsymbol{b_2}]$ does not explicitly depend on $t$.

## References

[1] L. D. Stone, C. M. Keller, T. M. Kratzke, and J. P. Strumpfer, "Search analysis for the underwater wreckage of Air France Flight 447," in *Proc. 14th Int. Conf. Inf. Fusion*, Jul. 2011, pp. 1–8.

[2] V. Schmidt, T. C. Weber, D. N. Wiley, and M. P. Johnson, "Underwater tracking of humpback whales (megaptera novaeangliae) with high-frequency pingers and acoustic recording tags," *IEEE J. Ocean. Eng.*, vol. 35, no. 4, pp. 821–836, Oct. 2010.

[3] N. A. Society and B. Amanda, *Underwater Archaeology: The NAS Guide to Principles and Practice*. Oxford, U.K.: Blackwell, 2009.

[4] M. D. Feezor, F. Y. Sorrell, P. R. Blankinship, and J. G. Bellingham, "Autonomous underwater vehicle homing/docking via electromagnetic guidance," *IEEE J. Ocean. Eng.*, vol. 26, no. 4, pp. 515–521, Oct. 2001.

[5] B. Ferreira, A. Matos, and N. Cruz, "Single beacon navigation: Localization and control of the MARES AUV," in *Proc. OCEANS Conf.*, 2010, pp. 1–9.

[6] P. Batista, C. Silvestre, and P. Oliveira, "Single range aided navigation and source localization: Observability and filter design," *Syst. Control Lett.*, vol. 60, no. 8, pp. 665–673, Aug. 2011.

[7] Y. T. Tan, R. Gao, and M. Chitre, "Cooperative path planning for range-only localization using a single moving beacon," *IEEE J. Ocean. Eng.*, vol. 39, no. 2, pp. 371–385, Apr. 2014.

[8] S. E. Webster, R. M. Eustice, H. Singh, and L. L. Whitcomb, "Advances in single-beacon one-way-travel-time acoustic navigation for underwater vehicles," *Int. J. Robot. Res.*, vol. 31, no. 8, pp. 935–950, Jul. 2012.

[9] B. Bayat, N. Crasta, H. Li, and A. Ijspeert, "Optimal search strategies for pollutant source localization," in *Proc. IEEE/RSJ Int. Conf. Intell. Robots Syst.*, Oct. 2016, pp. 1801–1807.

[10] G. Vallicrosa, J. Bosch, N. Palomeras, P. Ridao, M. Carreras, and N. Gracias, "Autonomous homing and docking for AUVs using range-only localization and light beacons," *IFAC-PapersOnLine*, vol. 49, no. 23, pp. 54–60, 2016.

[11] C. Zhang, D. Arnold, N. Ghods, A. Siranosian, and M. Krstic, "Source seeking with non-holonomic unicycle without position measurement and with tuning of forward velocity," *Syst. Control Lett.*, vol. 56, no. 3, pp. 245–252, Mar. 2007.

[12] N. Ghods and M. Krstic, "Source seeking with very slow or drifting sensors," *J. Dyn. Syst., Meas., Control*, vol. 133, no. 4, pp. 044–504, 2011.

[13] C. Zhang, A. Siranosian, and M. Krstic, "Extremum seeking for moderately unstable systems and for autonomous vehicle target tracking without position measurements," *Automatica*, vol. 43, no. 10, pp. 1832–1839, Oct. 2007.

[14] A. Scheinker and M. Krstic, "Extremum seeking with bounded update rates," *Syst. Control Lett.*, vol. 63, pp. 25–31, Jan. 2014.

[15] F. Mandić and N. Mišković, "Tracking underwater target using extremum seeking," *IFAC-PapersOnLine*, vol. 48, no. 2, pp. 149–154, 2015.

[16] M. Krstic and H.-H. Wang, "Stability of extremum seeking feedback for general nonlinear dynamic systems," *Automatica*, vol. 36, no. 4, pp. 595–601, 2000.

[17] F. Arrichiello, G. Antonelli, A. P. Aguiar, and A. Pascoal, "An observability metric for underwater vehicle localization using range measurements," *Sensors*, vol. 13, no. 12, pp. 16 191–16 215, 2013.

[18] N. Crasta, M. Bayat, A. P. Aguiar, and A. M. Pascoal, "Observability analysis of 2d single beacon navigation in the presence of constant currents for two classes of maneuvers," in *Control Appl. Mar. Syst.*, vol. 9, 2013, pp. 227–232.

[19] W. Cheng, A. Thaeler, X. Cheng, F. Liu, X. Lu, and Z. Lu, "Time-synchronization free localization in large scale underwater acoustic sensor networks," in *Proc. 29th IEEE Int. Conf. Distrib. Comput. Syst. Workshops*, 2009, pp. 80–87.

[20] A. Bensky, *Wireless Positioning Technologies and Applications*. Norwood, MA, USA: Artech House, 2007.

[21] F. Gustafsson and F. Gunnarsson, "Positioning using time-difference of arrival measurements," in *Proc. IEEE Int. Conf. Acoust., Speech, Signal Process.*, Apr. 2003, vol. 6, pp. VI–553.

[22] J. Smith and J. Abel, "The spherical interpolation method of source localization," *IEEE J. Ocean. Eng.*, vol. 12, no. 1, pp. 246–252, Jan. 1987.

[23] B. M. Ferreira, A. C. Matos, N. A. Cruz, and R. M. Almeida, "Towards cooperative localization of an acoustic pinger," in *Proc. OCEANS Conf.*, Oct. 2012, pp. 1–5.

[24] B. Kouzoundjian, F. Beaubois, S. Reboul, J. B. Choquel, and J. C. Noyer, "A TDOA underwater localization approach for shallow water environment," in *Proc. OCEANS Conf., Aberdeen, Scotland*, Jun. 2017, pp. 1–4.

[25] B. A. Francis and M. Maggiore, *Flocking and Rendezvous in Distributed Robotics*. Cham, Switzerland: Springer, 2016.

[26] D. R. Begault, *3D Sound for Virtual Reality and Multimedia*. San Diego, CA, USA: Academic, 1994.

[27] H.-B. Dürr, M. S. Stankovic, C. Ebenbauer, and K. H. Johansson, "Lie bracket approximation of extremum seeking systems," *Automatica*, vol. 49, pp. 1538–1552, 2013.

[28] N. Miskovic, D. Nad, N. Stilinovic, and Z. Vukic, "Guidance and control of an overactuated autonomous surface platform for diver tracking," in *Proc. 21st Mediterranean Conf. Control Autom.*, 2013, pp. 1280–1285.

[29] J. Neasham, "Towards large scale underwater communication networks—miniature, low cost, low power acoustic transceiver design." [Online]. Available: https://udrc.eng.ed.ac.uk/sites/udrc.eng.ed.ac.uk/files/attachments/Towards%20large%20scale%20underwater%20\\communication%20networks.pdf, Accessed: Sep. 20, 2018.

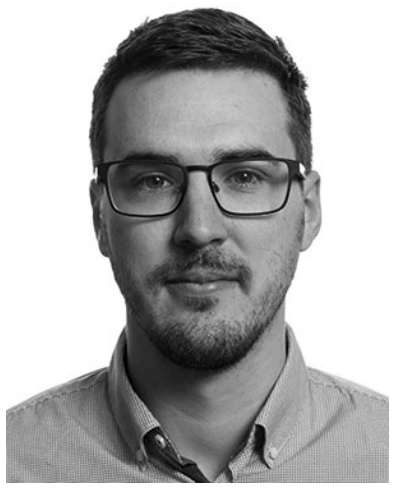

**Filip Mandić** (M'18) received the M.Sc. degree in underwater vehicle localization from the University of Zagreb in 2014, and is currently working toward the Ph.D. degree in control and localization of underwater vehicles at the Univerity of Zagreb, Zagreb, Croatia.

He is currently a Research and Teaching Assistant at the Faculty of Electrical Engineering and Computing, University of Zagreb. He is involved in European Union projects H2020 subCULTron, H2020 EXCELLABUST, and nationally funded HrZZ project CroMarX. His research interests include navigation and localization of underwater and surface marine vehicles, and extremum seeking control.

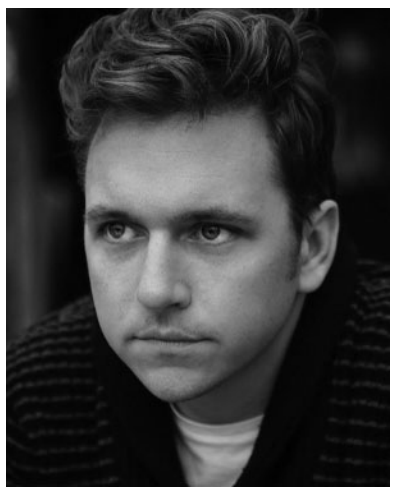

**Nikola Mišković** (S'05–M'08–SM'17) received the Ph.D. degree in control of marine vehicles from the University of Zagreb, Faculty of Electrical Engineering and Computing, in 2010.

He is currently an Associate Professor in Robotics and Control at the Faculty of Electrical Engineering and Computing, University of Zagreb, Zagreb, Croatia. He participated in 14 European projects (H2020, FP7, DG-ECHO, INTERREG), four Office of Naval Research Global (ONR-G) projects, two NATO projects, and seven national projects, out of which he coordinated seven. He has authored or coauthored more than 70 papers in journals and conference proceedings in the area of navigation, guidance and control, as well as cooperative control in marine robotics.

Dr. Mišković is the President of Chapter for Robotics and Automation of the Croatian Section from 2016 to 2019, IFAC (member of the Technical Committee on Marine Systems), and Centre for Underwater Systems and Technologies. In 2013, he was the recipient of the Young Scientist Award "Vera Johanides" of the Croatian Academy of Engineering (HATZ) for scientific achievements, and he was also the recipient of the Annual State Science Award for 2015, awarded by the Croatian Parliament.

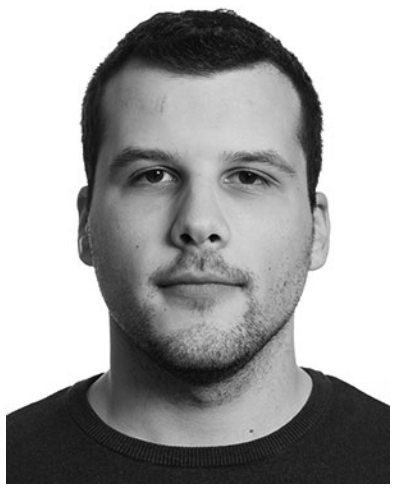

**Ivan Lončar** (S'19) received the M.Sc. degree in formation control of autonomous unmanned marine platforms from the University of Zagreb, Faculty of Electrical Engineering and Computing, in 2016. He is currently working toward the Ph.D. degree in marine robotics at the Faculty of Electrical Engineering and Computing, University of Zagreb, Zagreb, Croatia.

He is currently working as a Research and Teaching Assistant at the Faculty of Electrical Engineering and Computing, University of Zagreb. He is working on the European Union-funded H2020 subCULTron project, while collaborating on H2020 EXCELLABUST project, and nationally funded HrZZ project CroMarX. His research interest is navigation and localization of underwater and surface marine vehicles.